\documentclass[11pt]{article}

\usepackage{CJKutf8}
\usepackage{booktabs}

\usepackage{acl}

\usepackage{times}
\usepackage{latexsym}
\usepackage[T1]{fontenc}

\usepackage[utf8]{inputenc}
\usepackage{microtype}

\usepackage{inconsolata}

\usepackage{graphicx}
\usepackage{subcaption} 
\usepackage{amsmath}
\usepackage{amssymb}
\usepackage{bm}
\usepackage{float}
\usepackage{placeins} 
\usepackage{listings}
\usepackage{xcolor}
\usepackage{caption}
\lstdefinelanguage{json}{
    basicstyle=\ttfamily\small,
    numbers=none,
    numberstyle=\tiny\color{gray},
    stepnumber=1,
    numbersep=5pt,
    showstringspaces=false,
    breaklines=true,
    frame=none,
    backgroundcolor=\color{gray!10},
    literate=
     *{0}{{{\color{black}0}}}{1}
      {1}{{{\color{black}1}}}{1}
      {2}{{{\color{black}2}}}{1}
      {3}{{{\color{black}3}}}{1}
      {4}{{{\color{black}4}}}{1}
      {5}{{{\color{black}5}}}{1}
      {6}{{{\color{black}6}}}{1}
      {7}{{{\color{black}7}}}{1}
      {8}{{{\color{black}8}}}{1}
      {9}{{{\color{black}9}}}{1}
      {:}{{{\color{black}:}}}{1}
      {,}{{{\color{black},}}}{1}
      {"}{{{\color{red}"}}}{1},
}

\usepackage{multirow}
\usepackage{multicol}
\usepackage{array}

\usepackage[many]{tcolorbox} 
\definecolor{blue}{HTML}{5989cf}
\definecolor{sub}{HTML}{cde4ff}
\newtcolorbox{todobox}{
    colback = sub, 
    colframe = blue, 
    boxrule = 0pt, 
    leftrule = 6pt 
}
\usepackage{xspace}
\newcommand{\klar}{\textsc{KLAR}\xspace}

\title{Lost in Multilinguality: Dissecting Cross-lingual Factual Inconsistency\\in Transformer Language Models}

\author{
Mingyang Wang$^{1,2,3}$ \hspace*{0.2cm}
{Heike Adel$^{4}$} \hspace*{0.2cm} 
{Lukas Lange$^{1}$} \\ 
{\bf Yihong Liu$^{2,3}$} \hspace*{0.2cm} 
{\bf Ercong Nie$^{2,3}$} \hspace*{0.2cm} 
 {\bf Jannik Str\"{o}tgen$^{5}$ \hspace*{0.2cm} Hinrich Sch\"{u}tze$^{2,3}$} \\
  $^1$Bosch Center for Artificial Intelligence, Renningen, Germany \\
  $^2$LMU Munich, Germany \hspace*{0.2cm}
  $^3$Munich Center for Machine Learning (MCML) \\
  $^4$Hochschule der Medien, Stuttgart, Germany \\
  $^5$Karlsruhe University of Applied Sciences, Germany \\
  \texttt{mingyang.wang2@de.bosch.com} 
  }

\begin{document}
\maketitle
\begin{abstract}
Multilingual language models (MLMs) store factual knowledge across languages but often struggle to provide consistent responses to semantically equivalent prompts in different languages.
While previous studies point out this cross-lingual inconsistency issue, the underlying causes remain unexplored. In this work, we use mechanistic interpretability methods to investigate cross-lingual inconsistencies in MLMs. 
We find that MLMs encode knowledge in a language-independent concept space through most layers,
and only transition to language-specific spaces in the final layers. Failures during the language transition often result in incorrect predictions in the target language, even when the answers are correct in other languages.
To mitigate this inconsistency issue, we propose a linear shortcut method that bypasses computations in the final layers, enhancing both prediction accuracy and cross-lingual consistency. 
Our findings shed light on the internal mechanisms of MLMs and provide a lightweight, effective strategy for producing more consistent factual outputs.
\end{abstract}

\begin{figure}[t]
    \centering
    \includegraphics[width=1\linewidth]{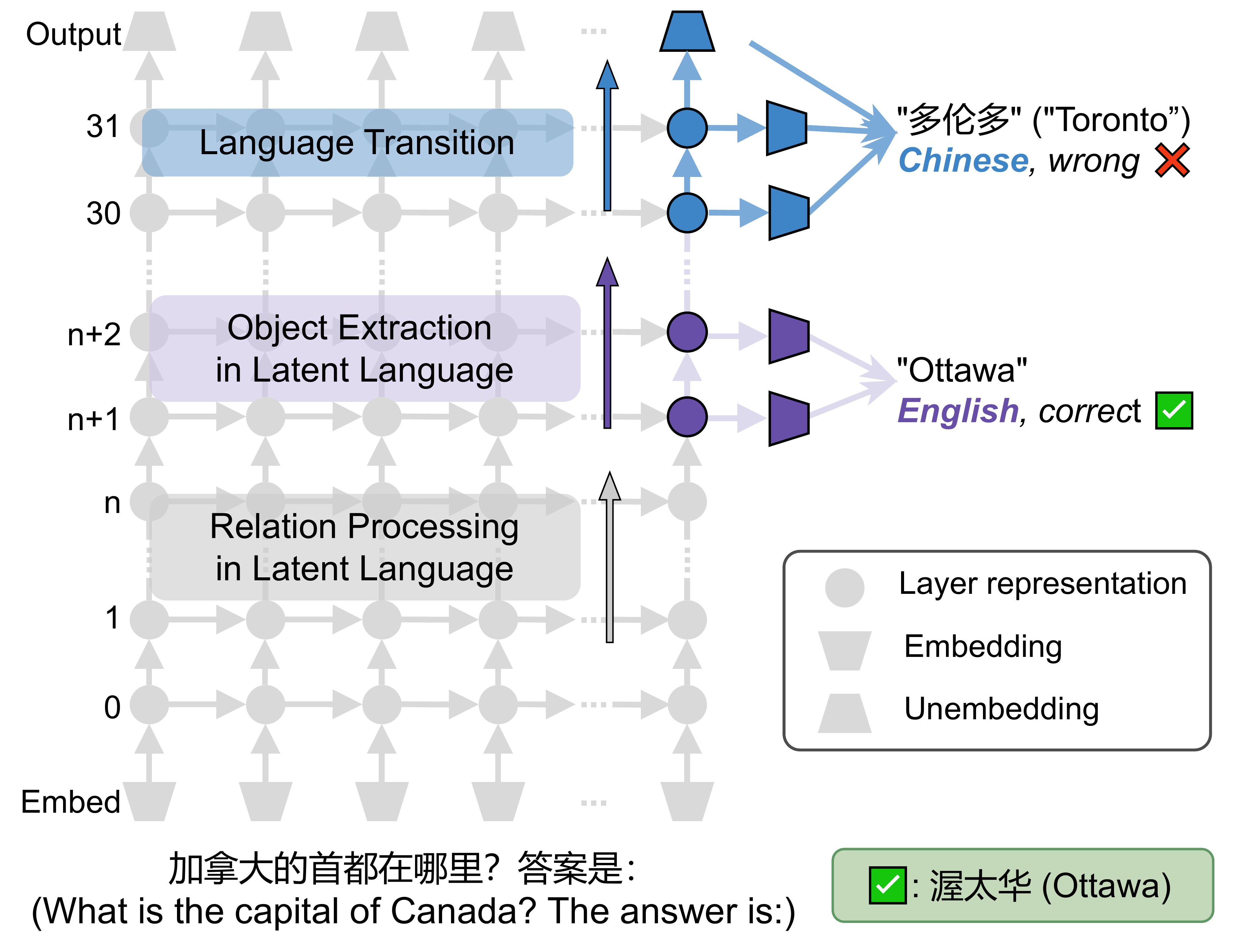}
   \caption[Caption for LOF]{Illustration of language transition failure in LLaMA2 when answering the question: 
   ``\begin{CJK}{UTF8}{gbsn}加拿大的首都在哪里?答案是：\end{CJK}'' (``What is the capital of Canada? The answer is:''). In intermediate layers, the model processes information in its latent language, i.e., a concept space independent of the input language.\protect\footnotemark\ While it correctly identifies ``Ottawa'' in English during the concept-space object extraction, the final output 
   ``\begin{CJK}{UTF8}{gbsn}多伦多\end{CJK}'' (``Toronto'') is incorrect after transitioning to Chinese. This indicates the model's failure to adapt knowledge from the concept space to the target language, leading to cross-lingual inconsistency.}
    \label{fig:teaser-image}
\end{figure}

\section{Introduction}

Multilingual language models (MLMs) have shown remarkable capabilities in storing and retrieving factual knowledge across languages \citep{jiang-etal-2020-x, kassner-etal-2021-multilingual}. However, they often exhibit inconsistencies when responding to semantically equivalent prompts in different languages. For instance, an MLM might correctly predict the capital of Canada when asked in English but fail to do so when queried in another language, e.g., Chinese. This phenomenon is known as \textit{cross-lingual factual inconsistency} \citep{qi-etal-2023-cross}. It raises questions about how effectively MLMs transfer knowledge across languages, and shows limitations in their robustness and fairness.

Understanding the root causes of such inconsistencies is crucial, yet research in this area remains limited. While prior studies have explored the inner workings of MLMs \citep{wendler-etal-2024-llamas, dumas2024llamas, fierro2024multilingual}, they mainly focus on scenarios where models make correct predictions, leaving the reasons behind inconsistent predictions unexplored. Furthermore, while \citet{qi-etal-2023-cross} identify frequent cross-language inconsistencies in MLMs, they do not investigate the underlying causes behind them.

\footnotetext{This concept space in LLaMA2, as seen through the Logit Lens \citep{logit-lens}, exhibits a bias towards English, reflecting its English-centric nature \citep{wendler-etal-2024-llamas}.}

In this work, we address this research gap by analyzing cross-lingual factual inconsistency through the lens of mechanistic interpretability \cite{olah2022mechanistic, nandaprogress}, which aims at reverse-engineering and, thereby, understanding language models. We trace information flows within MLMs to identify where inconsistencies arise on two complementary scenarios: (1) cases where models produce correct predictions consistent with English and (2) cases where models predicts correctly in English but generates incorrect answers in other languages.\footnote{English serves as the pivot language due to its central role in many multilingual language models \citep{held2023material, zhang-etal-2023-dont}.}
This comparison aims at uncovering the causes of both success and failure in multilingual factual recall.

Our analysis reveals that MLMs process factual knowledge in a concept space largely independent of the input language through most layers, and transition to language-specific spaces in the final layers. However, even when the correct prediction is encoded in this concept space, the model can fail the language transition, leading to incorrect predictions in the target language (see Figure \ref{fig:teaser-image}). This  highlights the critical role of the language transition mechanism for cross-lingual consistency.


Overall, our contributions are as follows: 

(\romannumeral1) \textbf{Dataset Construction (\textsection\ref{sec:dataset})}: We introduce \klar, an enhanced \textbf{K}now\textbf{L}edge probing dataset for \textbf{A}uto-\textbf{R}egressive models, covering 17 languages and 20 relation types. It provides a robust framework for  multilingual knowledge probing, which we use to evaluate the cross-lingual consistency of two state-of-the-art MLMs (\textsection\ref{sec:clc-evaluation}).

(\romannumeral2) \textbf{Mechanistic Analysis (\textsection\ref{sec:investigate-mechanisms})}: We conduct the first interpretability-driven study of cross-lingual factual inconsistency, revealing how MLMs encode and process factual knowledge across layers.

(\romannumeral3) \textbf{Failure Mode Identification (\textsection\ref{sec:examine-inconsistency})}: In a detailed layer-wise analysis, we identify the language transition mechanism as main failure point that leads to cross-lingual inconsistency.

(\romannumeral4) \textbf{Approach (\textsection\ref{sec:shortcut})}: We propose a shortcut method
that bypasses the model's final-layer computations, enhancing both prediction accuracy and cross-lingual consistency in MLMs.\footnote{Our data and code are open-source at \url{https://github.com/boschresearch/KLAR-CLC}}

\section{Related Work}


\paragraph{Mechanistic Interpretability (MI)} aims to understand LLMs by decomposing their computations into smaller, interpretable components. It has gained significant attention for studying factual knowledge recall in LLMs \citep{meng2022locating,dai-etal-2022-knowledge,geva-etal-2023-dissecting,yu-etal-2023-characterizing,lv2024interpreting,wang-etal-2024-unveiling, liu2025relation}. 

Following \citet{olah2020zoom} and \citet{rai2024practical}, MI research is categorized into the study of \textit{features}, which capture human-interpretable properties in model representations or components like neurons and attention heads \citep{elhage2022solu, gurnee2023finding}, and the study of \textit{circuits}, which refer to subgraphs of the model's computation graph responsible for implementing specific behaviors \citep{wang2022interpretability, elhage2021mathematical}.

In this work, we focus on representation-level feature-based interpretability analysis to interpret the behavior of multilingual language models in the knowledge probing task. 
Specifically, we use Logit Lens \citep{logit-lens} to project latent state representations of LMs into the vocabulary space, enabling the analysis of intermediate representations and tracking how information evolves across layers.



\paragraph{Interpreting Multilingual Language Models.}
Recent studies have explored the internal workings of MLMs. \citet{wendler-etal-2024-llamas} examine the latent language of LLaMA2 models using controlled translation, completion, and cloze tasks, finding that LLaMA2 internally relies on English as a pivot language. Building on this setup, \citet{dumas2024llamas} investigate the disentanglement of language and concept representations, demonstrating that LLaMA2 processes language and concept information independently. \citet{fierro2024multilingual} analyze knowledge probing tasks to study how mechanisms identified in monolingual contexts generalize to multilingual settings, but their focus remains limited to correct prediction cases.  

In contrast, our work centers on understanding the internal mechanisms responsible for cross-lingual inconsistencies. By examining both consistent and inconsistent predictions, we uncover how MLMs transition from language-independent to language-specific processing. This approach offers new insights into how MLMs encode and transfer factual knowledge across languages, addressing a key gap in prior research.

\section{\klar Dataset}
\label{sec:dataset}
We focus on the factual knowledge probing task, where a fact is represented as a subject-relation-object triple $\langle s_i, r_i, o_i \rangle$ and expressed in natural language prompts. Given a prompt constructed from the subject $s_i$ and relation $r_i$, LMs are expected to predict the object $o_i$. For example, the fact 
$\langle$\textit{Canada, capital, Ottawa}$\rangle$ 
can be queried as, ``What is the \textit{capital} of \textit{Canada}?'', and the model should predict the object \textit{Ottawa} as the answer.

\citet{qi-etal-2023-cross} introduce the BMLAMA17 dataset for evaluating multilingual factual knowledge in MLMs. However, in many factual questions in BMLAMA17, the object appears in the middle of the sentence rather than at the end, which is incompatible with knowledge probing for auto-regressive models. Furthermore, BMLAMA17 includes many relations with multiple correct answers,\footnote{For example, the relation "shares\_border\_with" (prompt: "Which country does <subject> share a border with?") often involves multiple correct answers, as a country typically shares borders with several others.} making it difficult to reliably evaluate the correctness of a model's response for a given $\langle s_i, r_i, o_i \rangle$ triple where $o_i$ is only one of the possible answers.

To address these limitations, we construct \klar, a \textbf{K}now\textbf{L}edge probing dataset that ensures compatibility with \textbf{A}uto-\textbf{R}egressive models and provides clarity in factual evaluation. We extract parallel factual knowledge triples in 17 languages from BMLAMA17 and design prompts where the object consistently appears at the end. Relation-specific templates are structured as ``<\textit{Question}> The answer is:'', e.g., $\langle$\textit{Canada, capital, Ottawa}$\rangle$ becomes: ``What is the capital of Canada? The answer is:''. These templates are initially created in English and translated into 16 other languages using \texttt{gpt-3.5-turbo}.
To ensure clarity, we exclude relations with multiple correct answers and inspect the semantic clarity in prompt templates manually and/or through back-translation.

The resulting \klar dataset includes 2,619 parallel factual knowledge triples in 17 languages, covering 20 relation types. Table~\ref{tab:klar-stats} provides an overview of the languages and sample relations. Detailed statistics are provided in Appendix~\ref{sec:appendix-dataset}.


\begin{table}[h]
\footnotesize
  \centering
    \begin{tabular}{lp{4.5cm}} 
    \midrule
    \textbf{Languages (17)} \\ 
    \midrule
    \multicolumn{2}{p{7cm}}{
    Arabic (\textit{ar}), Catalan (\textit{ca}), Greek (\textit{el}), English (\textit{en}), Spanish (\textit{es}), Persian (\textit{fa}), French (\textit{fr}), Hebrew (\textit{he}), Hungarian (\textit{hu}), Japanese (\textit{ja}), Korean (\textit{ko}), Dutch (\textit{nl}), Russian (\textit{ru}), Turkish (\textit{tr}), Ukrainian (\textit{uk}), Vietnamese (\textit{vi}), Chinese (\textit{zh})
    } \\
    \midrule
    \textbf{Relations (4/20)} & \textbf{Prompt example} \\
    \midrule
    capital & What is the capital of <subject>? The answer is: \\
    continent & Which continent is <subject> located in? The answer is: \\
    field\_of\_work & What field does <subject> work in? The answer is:  \\
    religion & What is the religious belief of <subject>? The answer is: \\
    \midrule
    \end{tabular}%
  \caption{Overview of the languages and 4 sample relations (out of 20 relations in total) in \klar.}
  \label{tab:klar-stats}%
\end{table}%

\section{Cross-lingual Consistency Evaluation}
\label{sec:clc-evaluation}

\paragraph{Models and Languages}
We analyze two widely used open-source multilingual auto-regressive language models: LLaMA2-7B \citep{touvron2023llama} and BLOOM-560M \citep{le2023bloom}. LLaMA2 is trained on a multilingual corpus dominated by English, which accounts for 89.7\% of the data, whereas BLOOM's training data is more balanced, with English comprising 31.3\% of the corpus.
Our analysis considers the languages shared between each model and our dataset, covering 12 languages for LLaMA2 and 7 for BLOOM. Details on the selected languages are provided in Table~\ref{tab:klar-languages-all} in Appendix~\ref{sec:appendix-dataset}.

\paragraph{Evaluation}
Many prior studies \citep{geva-etal-2023-dissecting,
  qi-etal-2023-cross, hernandezlinearity} assess
correctness based on the model's first predicted token. However, this approach is problematic,
especially in multilingual settings with complex
tokenization. In many cases, even if the model predicts the
correct first token, its complete output can still be
incorrect.\footnote{For example, given the Chinese prompt ``\begin{CJK}{UTF8}{gbsn}文森山位于哪个大陆？答案是：\end{CJK}''
(``Which continent is Vinson Massif located in? The answer is:''), the
BLOOM model outputs ``\begin{CJK}{UTF8}{gbsn}南美洲\end{CJK}'' (``South America'') instead of the
correct answer ``\begin{CJK}{UTF8}{gbsn}南极洲\end{CJK}''
(``Antarctica''). Although both
responses share the same first token, the final prediction
is incorrect.} To address this issue, we evaluate
correctness based on the model's full answer to each factual
question rather than relying solely on the first token.
Following \citet{jiang-etal-2020-x}, we evaluate cross-lingual consistency using the overlap ratio of correct predictions for parallel facts between language pairs.\footnote{We do not adopt the candidate-based consistency metric proposed by \citet{qi-etal-2023-cross}, as it relies on the next-token prediction, which, as discussed in Section~\ref{sec:clc-evaluation}, is unreliable in a multilingual setup.}

\begin{figure}[t]
    \centering
    \includegraphics[width=0.75\linewidth]{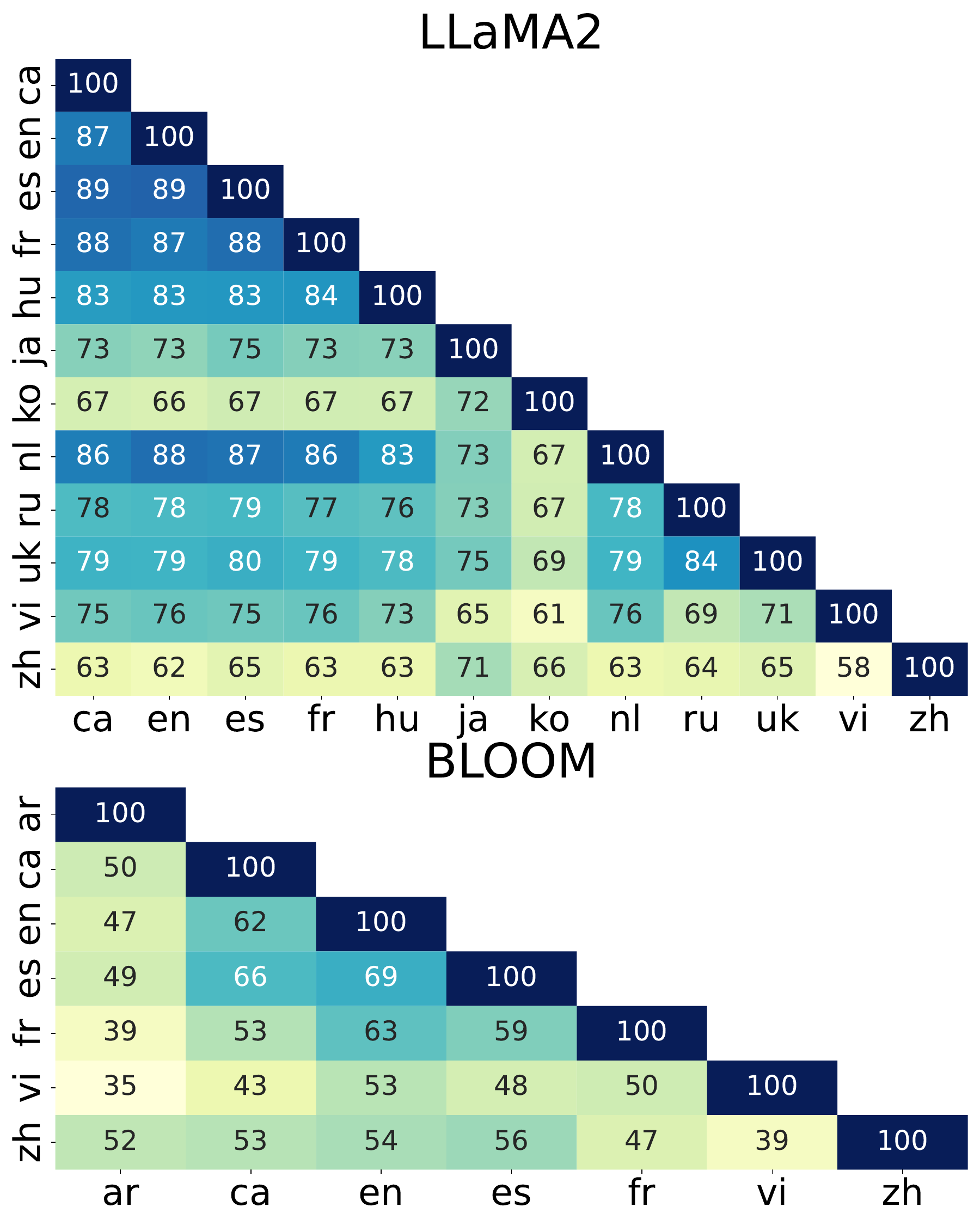}
    \caption{Cross-lingual consistency results across language pairs. The heatmaps show the overlap ratio of correct predictions between language pairs.}
    \label{fig:clc-results}
\end{figure}

\paragraph{Results}
\label{sec:measure-clc}

Figure~\ref{fig:clc-results} shows the cross-lingual consistency results for LLaMA2 and BLOOM. While LLaMA2 generally performs better than BLOOM, both models face challenges in achieving high consistency across languages, particularly between linguistically diverse pairs. The impact of language scripts is especially evident: Non-Latin scripts, such as Arabic (\textit{ar}), Chinese (\textit{zh}), and Korean (\textit{ko}), consistently show lower consistency scores. This underscores that cross-lingual consistency remains a key limitation for both models, emphasizing the need for more robust approaches to effectively analyze and address this issue.

\begin{figure*}[h!]
    \centering

    \begin{subfigure}{\linewidth}
        \centering
        \includegraphics[width=0.45\linewidth]{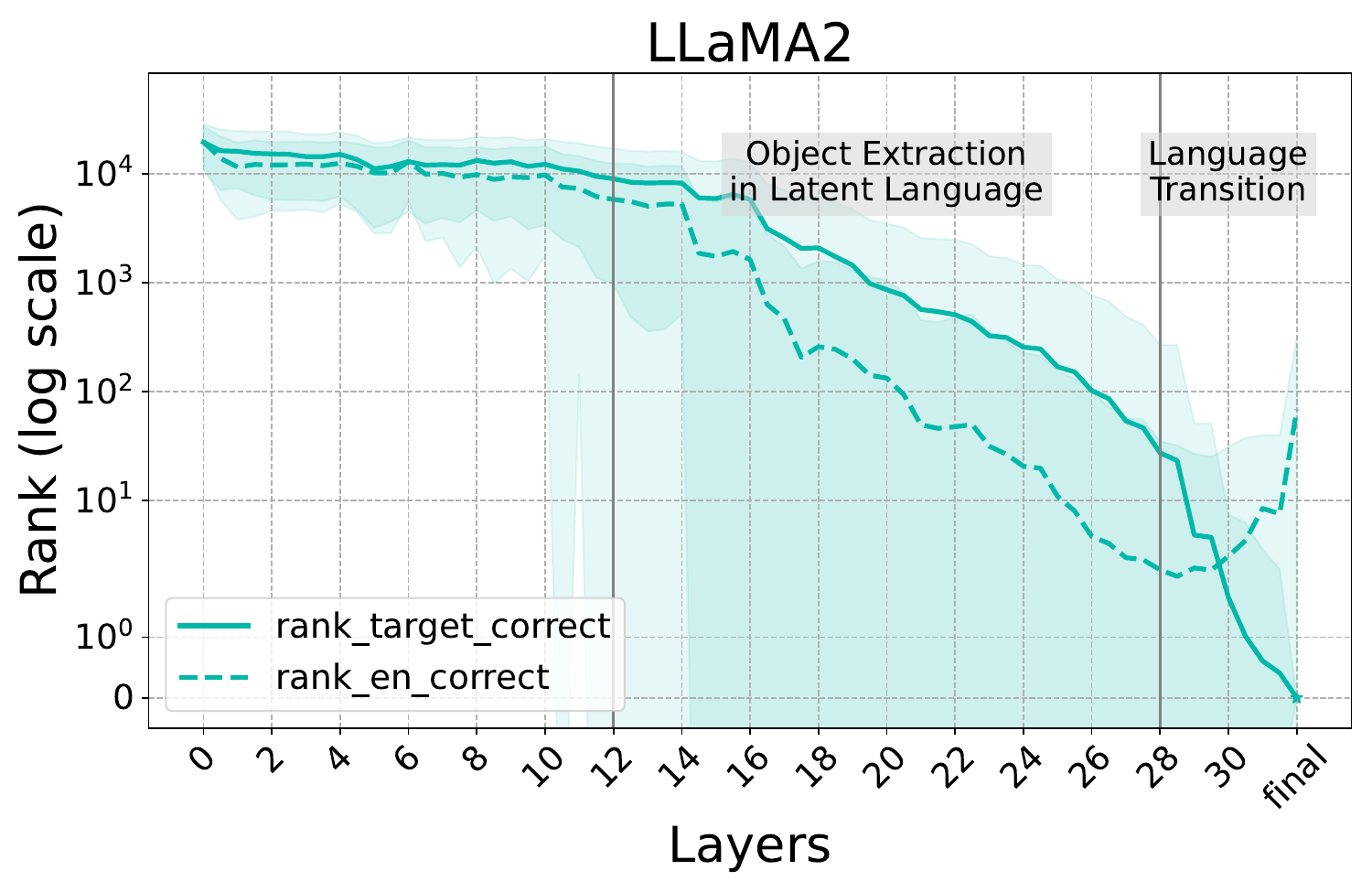}
        \hfill
        \includegraphics[width=0.45\linewidth]{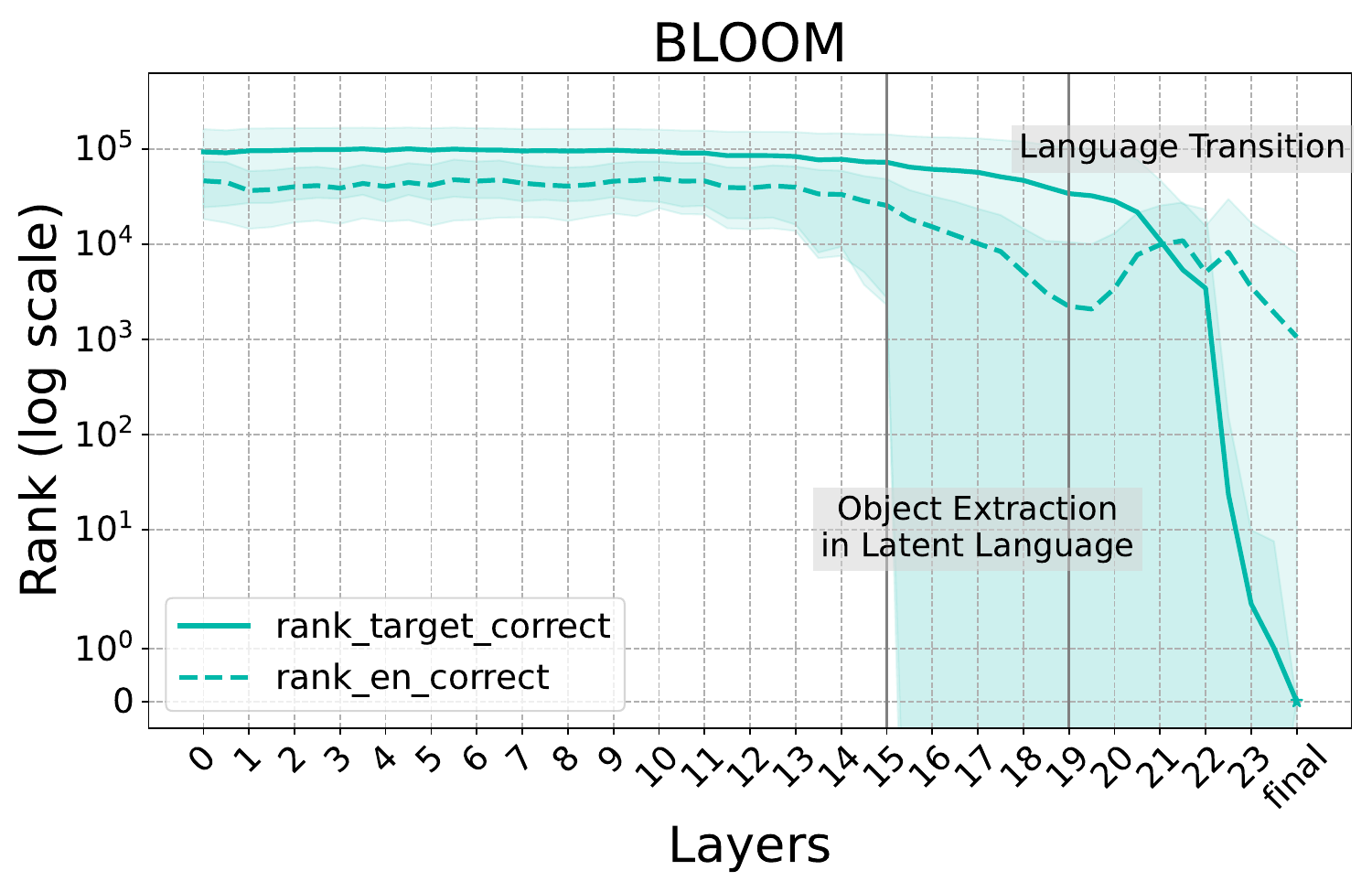}
        \caption{Layer-wise rank of correct predictions averaged across all languages and relations (\textsection \ref{subsec:investigate-rank}). ``\texttt{rank\_target\_correct}'' denotes the rank of correct predictions in the target language, while ``\texttt{rank\_en\_correct}'' represents the rank of their English equivalents.}
        \label{fig:mechanism-row1}
    \end{subfigure}

    \begin{subfigure}{\linewidth}
        \centering
        \includegraphics[width=0.45\linewidth]{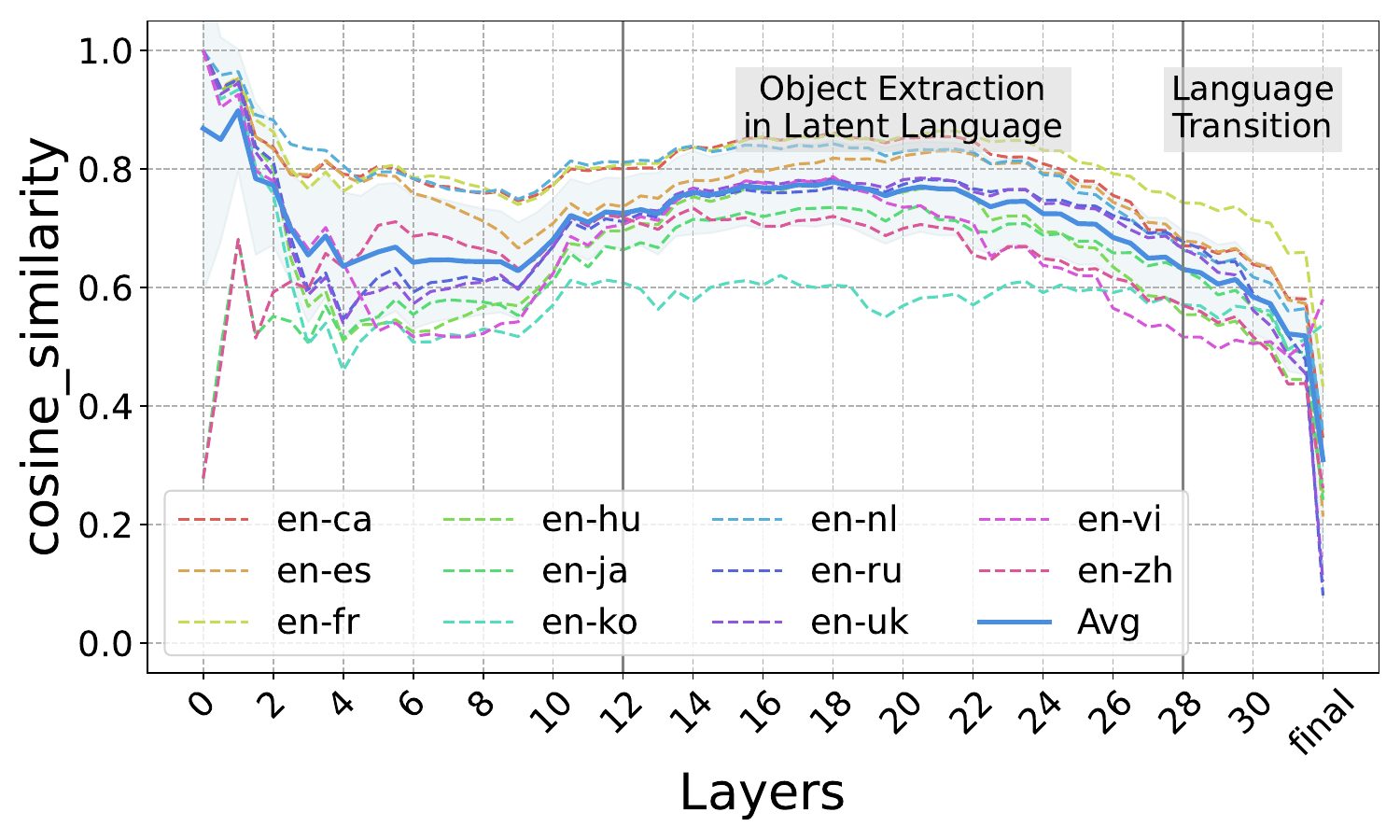}
        \hfill
        \includegraphics[width=0.45\linewidth]{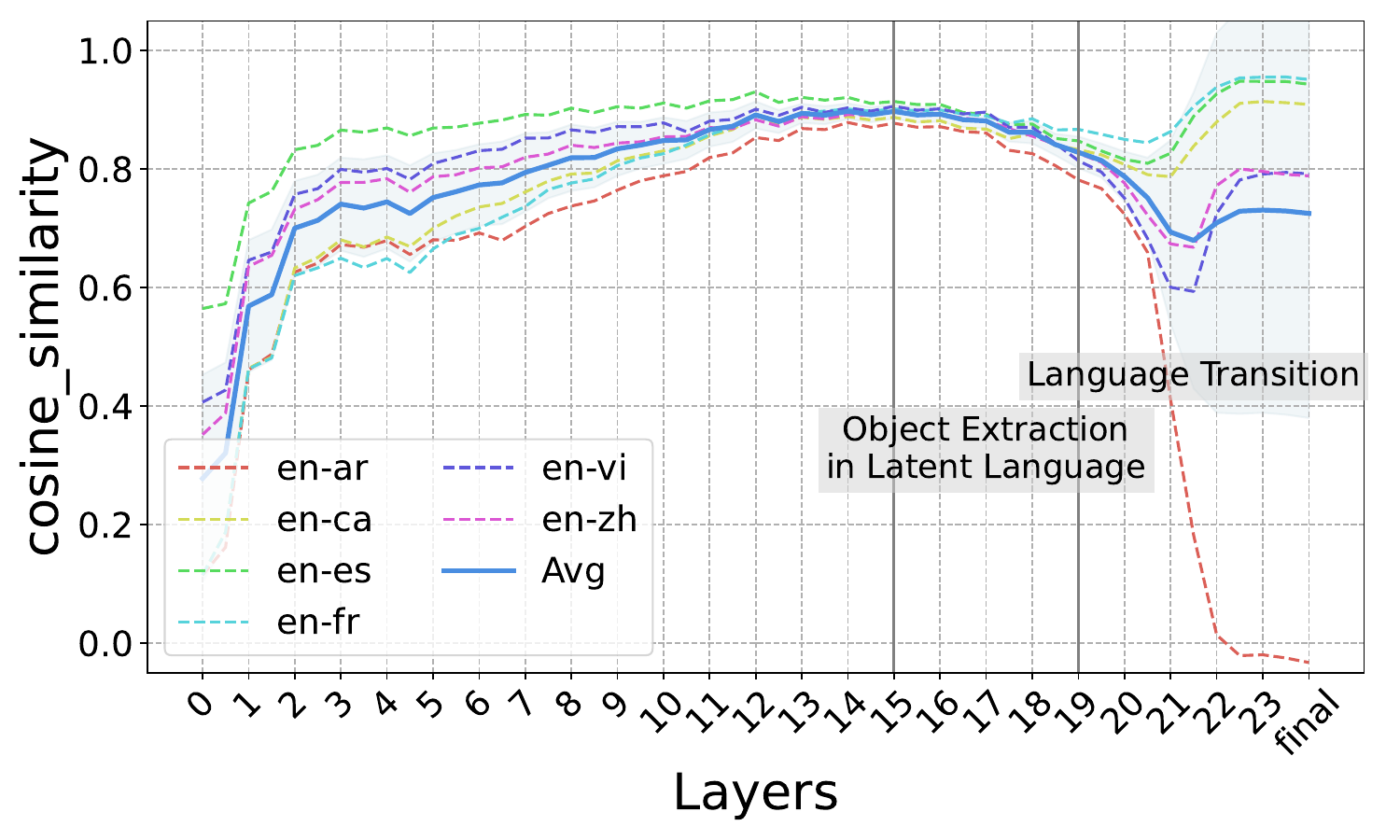}
        \caption{Cosine similarity of latent state similarity between each language pair averaged across all relations (\textsection \ref{subsec:investigate-hidden}).}
        \label{fig:mechanism-row2}
    \end{subfigure}

    \begin{subfigure}{\linewidth}
        \centering
        \includegraphics[width=0.45\linewidth]{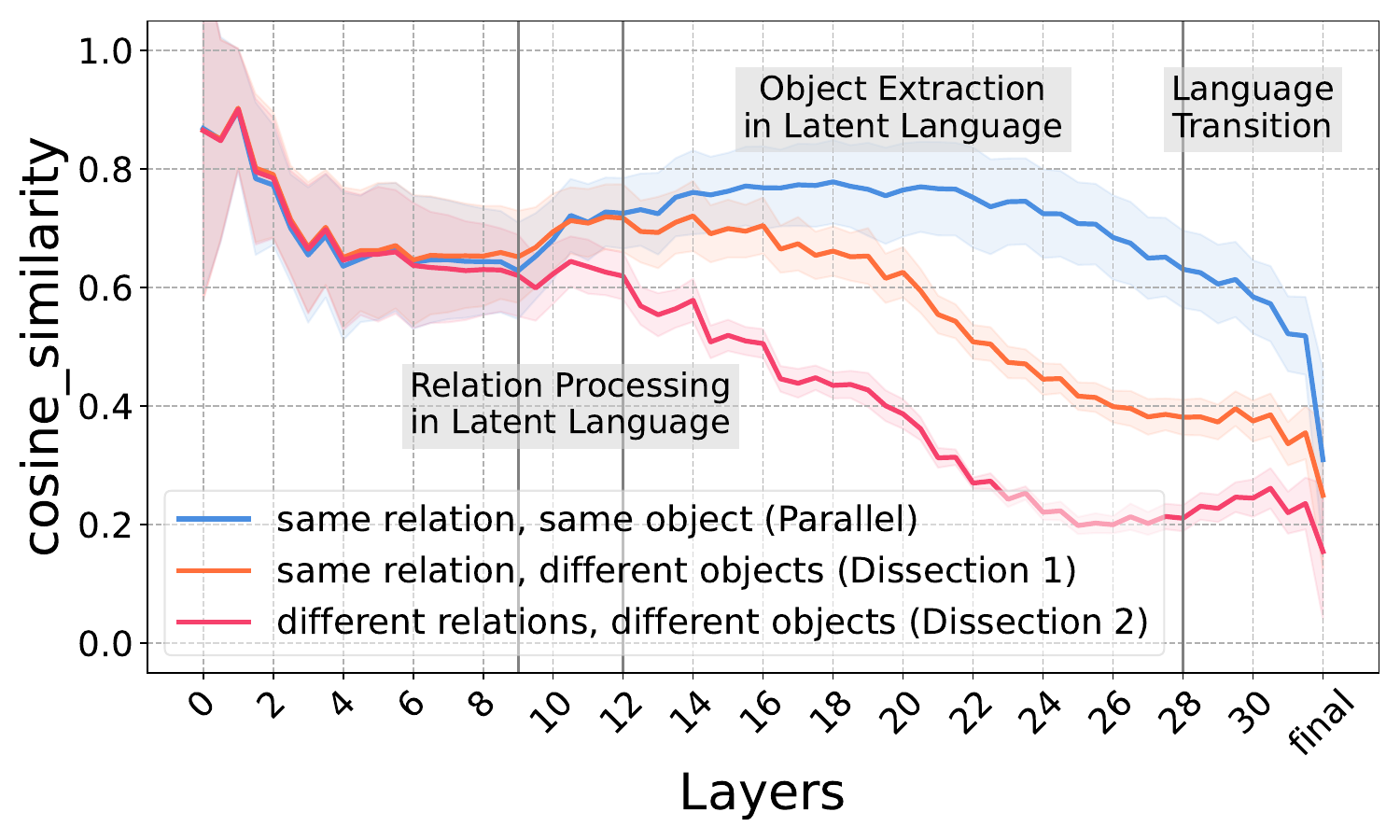}
        \hfill
        \includegraphics[width=0.45\linewidth]{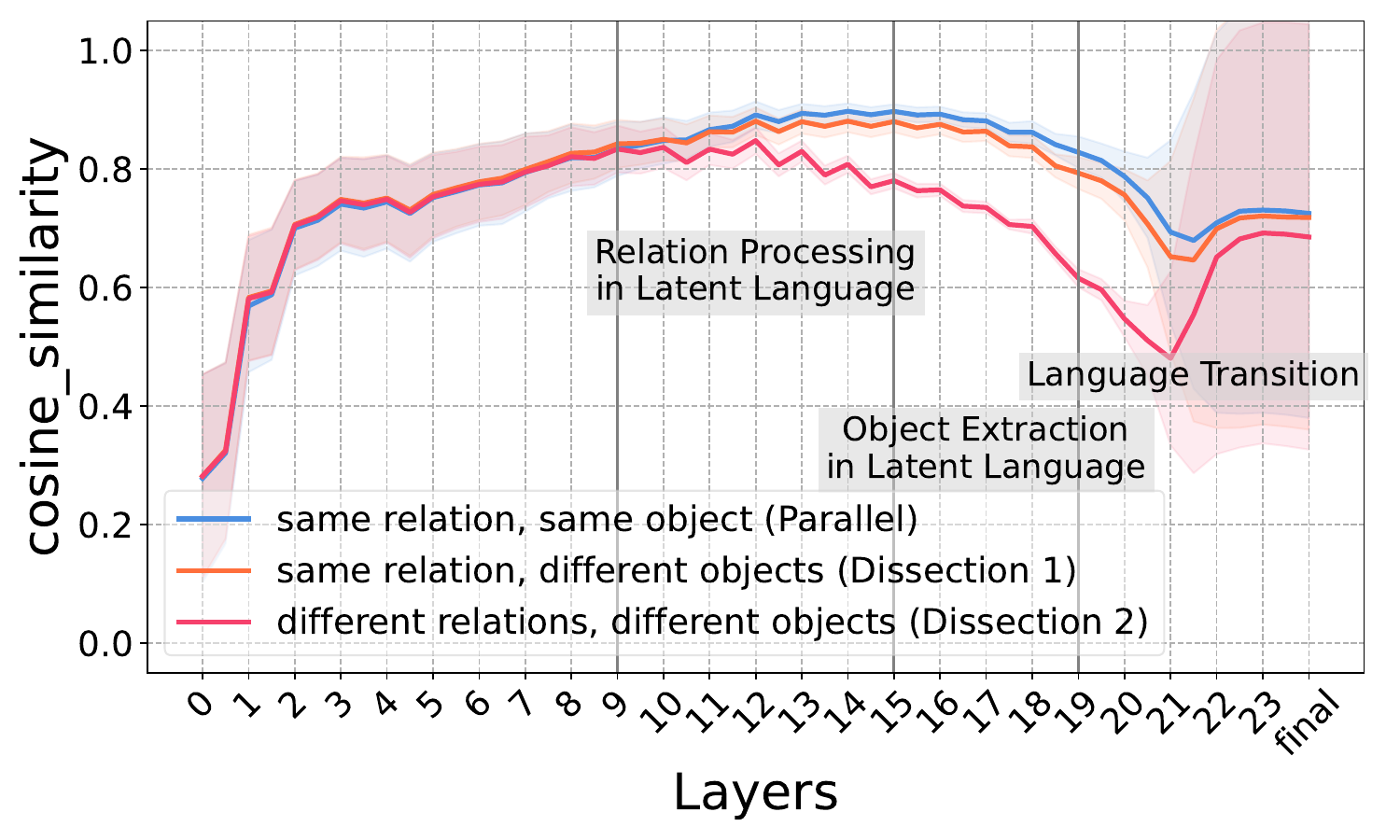}
        \caption{Comparative study of latent state similarity across language pairs (\textsection \ref{subsec:investigate-ablation}). We compare the latent state similarity for parallel facts, non-parallel facts sharing the same relation, and non-parallel facts belonging to different relations, respectively.}
        \label{fig:mechanism-row3}
    \end{subfigure}

    \caption{Analysis of multilingual knowledge probing of LLaMA2 and BLOOM, including (\ref{fig:mechanism-row1}) layer-wise evolution of correct prediction ranks, (\ref{fig:mechanism-row2}) latent state similarities across languages, and (\ref{fig:mechanism-row3}) the development of latent state similarities  in different settings.}
    \label{fig:mechanism-correct}
\end{figure*}

\section{Analyzing Multilingual Factual Recall}
\label{sec:investigate-mechanisms}
To understand how multilingual language models recall factual knowledge across languages, we analyze their internal mechanisms from multiple perspectives: the layer-wise evolution of prediction ranks (\textsection \ref{subsec:investigate-rank}), latent state similarities across language pairs (\textsection \ref{subsec:investigate-hidden}), information flow within the model (\textsection \ref{subsec:investigate-ablation}), and the composition of the latent concept space (\textsection \ref{subsec:language-composition}). 

\subsection{From the Perspective of Rankings}
\label{subsec:investigate-rank}

First, we use Logit Lens \cite{logit-lens} to project latent states at each layer to the vocabulary (unembedding) and measure the rank (the lower, the better) of the target object at each layer. Specifically, we compare the rank of the correct object in its target language (\texttt{rank\_target\_correct}) and its English equivalent (\texttt{rank\_en\_correct}). This approach allows us to trace how the model processes factual knowledge across layers and transitions between different representation modes.

Figure~\ref{fig:mechanism-row1} shows distinct phases of knowledge processing in both models. In the early layers, both ranks remain high, indicating that the models have not begun extracting the target object. Around layer 15 in BLOOM and layer 12 in LLaMA2, both (\texttt{rank\_target\_correct}) and (\texttt{rank\_en\_correct}) drop significantly, marking the beginning of the object extraction phase.

This phase continues until layer 28 in LLaMA2 and layer 19 in BLOOM, where a notable divergence occurs. The English rank (\texttt{rank\_en\_correct}) begins to increase, while the target-language rank (\texttt{rank\_target\_correct}) continues to decrease. This divergence reflects a transition from language-independent object extraction to target language-specific object extraction, where the models adapt the representations to align with the target language.

These findings show that MLMs recall know\-ledge through an initial concept-space object extraction phase (marked by significant rank drops for both English and target language answers) before transitioning to language-specific object extraction and producing the final output.

\subsection{From the Perspective of Latent States}
\label{subsec:investigate-hidden}
Moreover, we measure the cosine similarity of latent states between language pairs across layers.

Figure~\ref{fig:mechanism-row2} shows the average cosine similarity of latent states between English and individual target languages for LLaMA2 and BLOOM.\footnote{For clarity, only language pairs involving English are shown here. Complete results for all language pairs are provided in Appendix~\ref{sec:appendix-similarity}.} As information propagates through the layers, similarity increases, peaking around 0.8 in the middle layers for both models.\footnote{Our similarity analysis focuses on the final token, and all prompts end with "The answer is:". In LLaMA2, the colon “:” is typically tokenized as a standalone final token, leading to high early-layer similarity—except in en-ja and en-zh, which use language-specific colon variants. In contrast, BLOOM often fuses the colon with the preceding word (e.g., "is:", "es:", "là:", "\begin{CJK}{UTF8}{gbsn}是：\end{CJK}") or tokenizes it separately (e.g., in ar, ca, fr), causing lower similarity due to mismatched token boundaries. This pattern is also visible in Figure 9.} This trend holds even for linguistically diverse pairs, such as English and Arabic, suggesting the formation of a shared concept space where factual knowledge is encoded in the model's latent language which is generic and independent of the input language.
In the final layers, similarity decreases, reflecting a transition to language-specific processing. This aligns with the divergence observed in Section~\ref{subsec:investigate-rank}, where the rank changes of the target language object and its English equivalent begin to differ. These observations confirm the model's transition from concept-space object extraction to language-specific adaptations in the final layers.

\subsection{Information Flow Dissection}
\label{subsec:investigate-ablation}

While Sections~\ref{subsec:investigate-rank} and \ref{subsec:investigate-hidden} demonstrate the presence of a concept space in the middle layers, they do not clarify the type of information contributing to the observed high similarity between language pairs. To disentangle whether this similarity arises from relational information, object information, or both, we perform comparative experiments under three conditions: (1) \textbf{Same relation, same object} (\textbf{\textit{Parallel}}, as in Section~\ref{subsec:investigate-hidden}): Latent state similarity is calculated using parallel facts between each language pair (e.g., "the capital of Canada" in both English and another language);
(2) \textbf{Same relation, different objects (\textit{Dissection 1})}: Similarity is calculated using non-parallel facts sharing the same relation (e.g., "the capital of Canada" in one language versus "the capital of Spain" in another);
(3) \textbf{Different relation, different objects (\textit{Dissection 2})}: Similarity is calculated using non-parallel facts from different relations (e.g., "the capital of Canada" versus "the official language of Spain").

Figure~\ref{fig:mechanism-row3} shows distinct processing phases. Around layer 9, the \textit{Dissection 2} curve drops significantly in both models, while \textit{Parallel} and \textit{Dissection 1} curves remain close, indicating that models process relational information specific to the current fact's relation. The high similarity during this stage suggests that such relation processing happens in a language-independent concept space.

From layer 12 in LLaMA2 and layer 15 in BLOOM, the \textit{Dissection 1} curve begins to drop, marking a transition to object-specific processing. During layers 12--28 in LLaMA2 and layers 15--19 in BLOOM, the \textit{Parallel} curve remains high, indicating that object information is processed in the model's latent language.

At layer 28 in LLaMA2 and layer 19 in BLOOM, the \textit{Parallel} curve drops significantly, signaling the language transition phase, where the concept-space object representations are adapted to the target language. 

Together, the progression shows the models' transitions from relation processing to object extraction and to language-specific adaptation.

\subsection{Concept Space Language Composition}
\label{subsec:language-composition}

To further explore how the concept space encodes information in MLMs, we analyze the language composition of their latent states. Using Logit Lens, we project intermediate layer representations onto the vocabulary space and identify the language of the top-10 predicted tokens at each layer using fasttext \citep{joulin-etal-2017-bag}.\footnote{We filter out tokens with confidence scores below 0.5. 
}
 
\begin{figure}[h]
    \centering
    \includegraphics[width=0.9\linewidth]{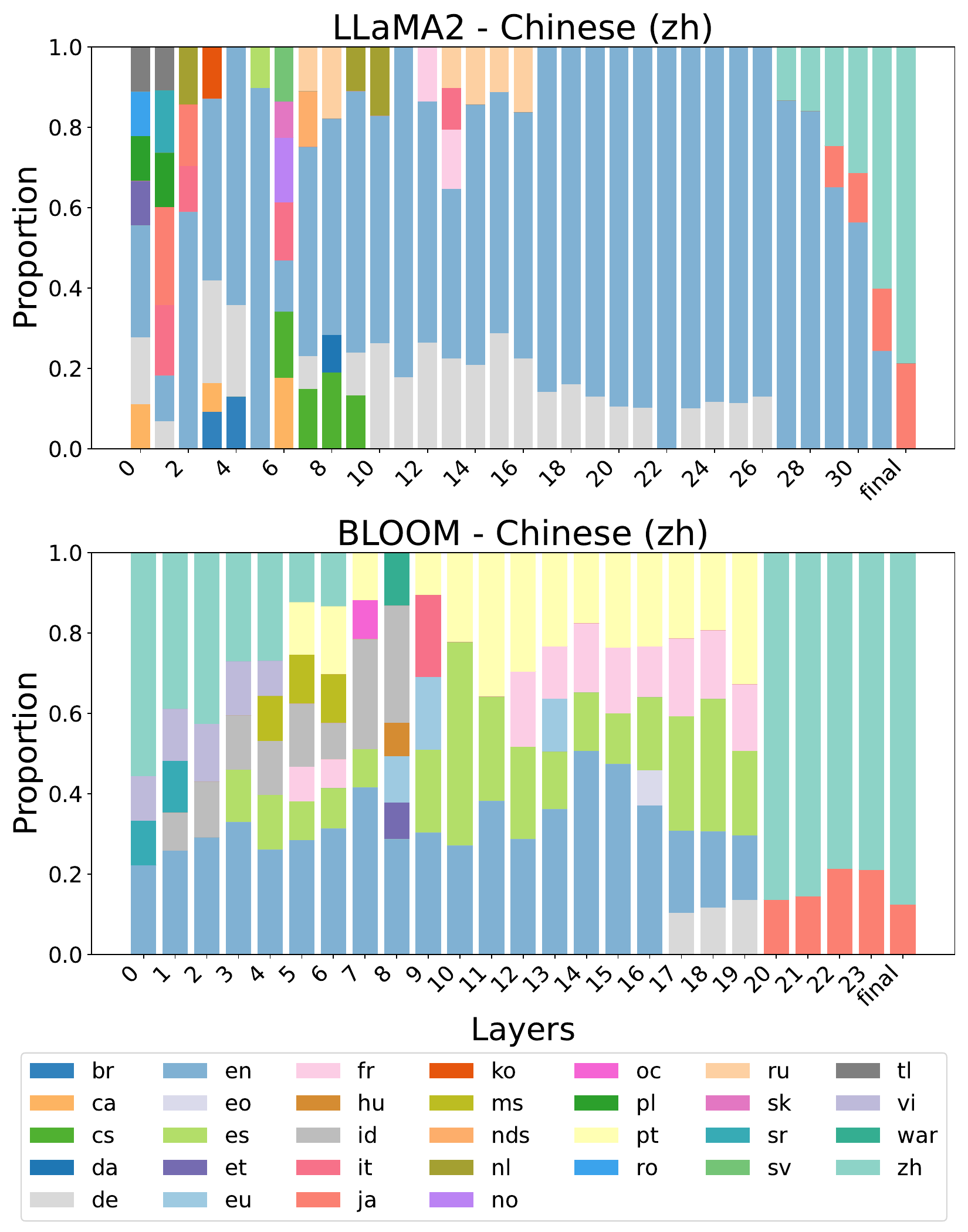}
    \caption{Language composition of latent
      representations with Chinese as the input language. 
      In LLaMA2, English dominates the middle-to-upper layers, whereas BLOOM has a more diverse language composition.}
    \label{fig:language-composition-zh}
\end{figure}

Figure~\ref{fig:language-composition-zh} shows the language composition for LLaMA2 and BLOOM with Chinese (zh) as the input language, averaged across factual queries spanning all relations. Results for other input languages are provided in Appendix~\ref{sec:appendix-results-rank-predictions}.

In LLaMA2, English dominates the middle-to-upper layers, suggesting that factual knowledge is processed in an English-centric concept space. This is consistent with prior findings that ``LLaMA2 models think in English'' \citep{wendler-etal-2024-llamas}. In contrast, BLOOM shows a more diverse composition in the middle-to-upper layers, comprising primarily Latin-based languages like English, French, Spanish, German, etc. 

Within each model, the middle-to-upper layers exhibits similar language compositions across different input languages (see Appendix Figures~\ref{fig:language-composition-shared} and \ref{fig:language-composition-unique}). This suggests that multilingual models encode factual knowledge in a shared concept space largely independent of the input language. Notably, this space is not necessarily aligned with any single language, indicating that \textbf{multilingual LLMs "think" in their own concept space} rather than in the surface form of a particular language.

\subsection{Summary}
\label{subsec:summary}
Our analysis reveals a three-stage knowledge recall process in MLMs (as illustrated in Figure~\ref{fig:teaser-image}): first relation processing, then object extraction in the model's latent language,
and finally the transition to language-specific processing to adapt the object to the target language. These findings provide a comprehensive view on the mechanisms of multilingual factual recall.


\begin{figure}[h]
    \centering
    \begin{subfigure}{\linewidth}
        \centering
        \includegraphics[width=0.9\linewidth]{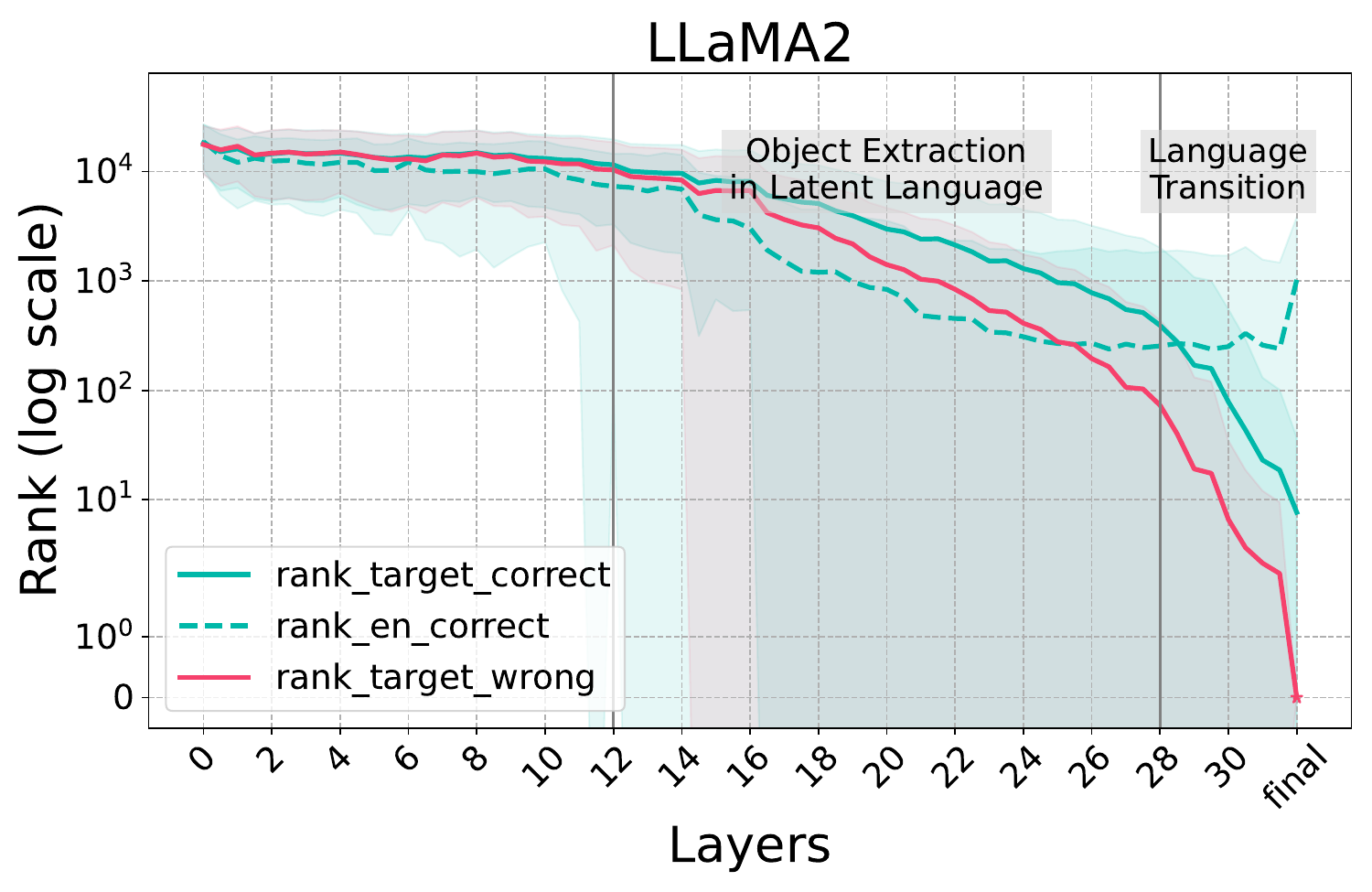}
        \includegraphics[width=0.9\linewidth]{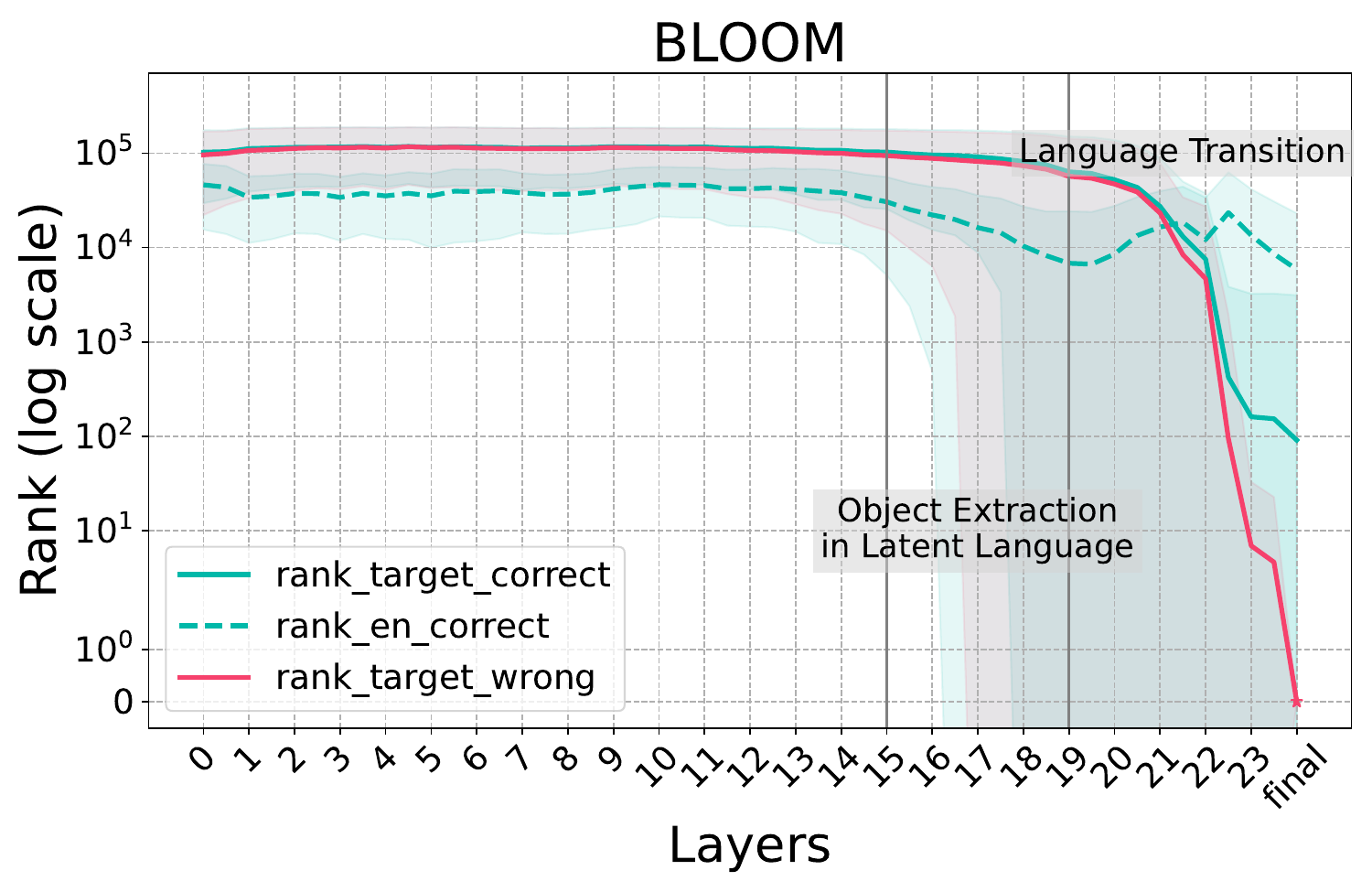}
    \end{subfigure}

    \caption{Layer-wise rank of incorrect predictions averaged across all languages and relations. The \texttt{rank\_target\_wrong} curve represents the rank of the model's final incorrect prediction across layers, while \texttt{rank\_target\_correct} and \texttt{rank\_en\_correct} denote the ranks of the correct answer in the target language and the English equivalent, respectively. 
    }
    \label{fig:mechanism-incorrect}
\end{figure}

\begin{figure}[h!]
    \centering
    \begin{subfigure}{\linewidth}
        \centering
        \includegraphics[width=0.9\linewidth]{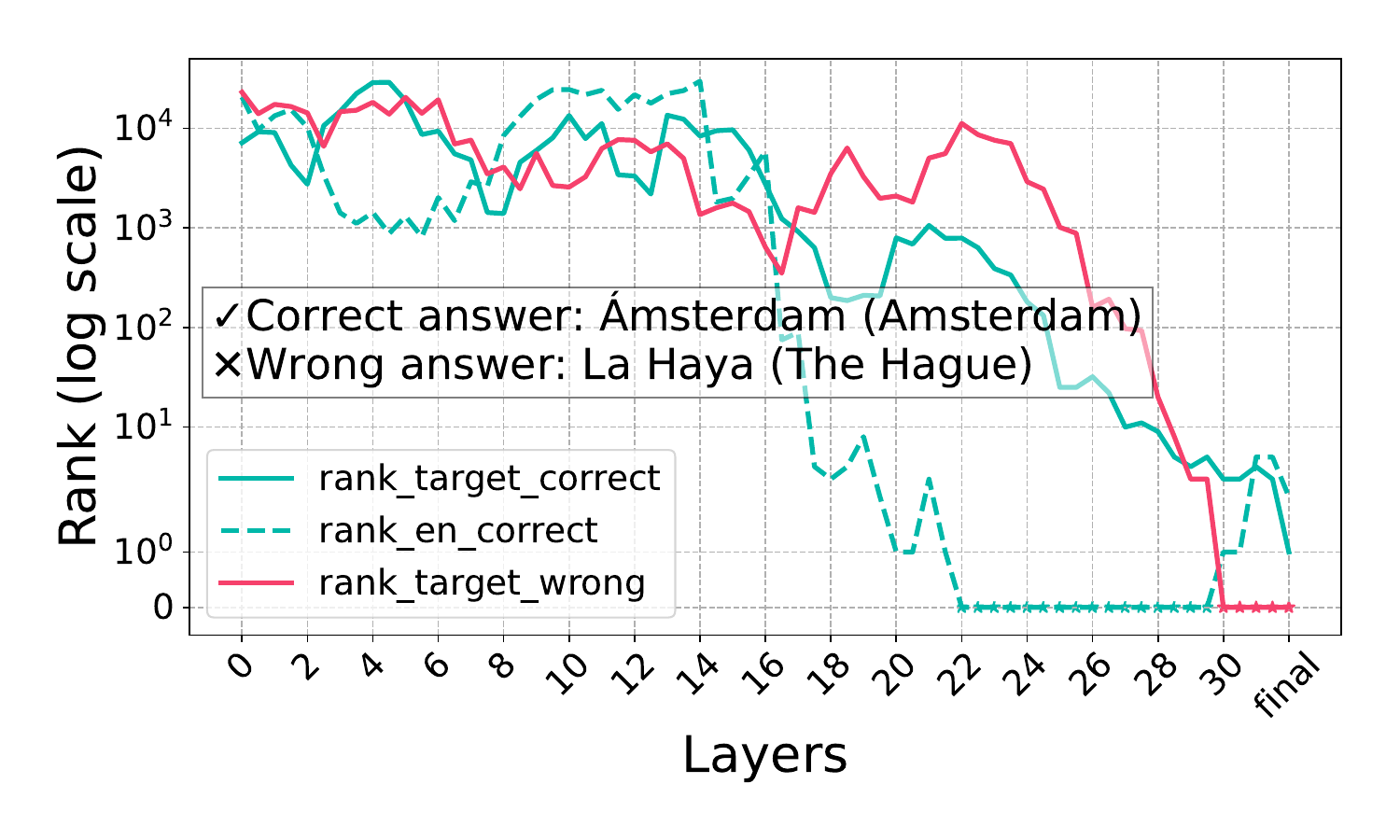}
        \caption{Prompt in Spanish:``\textquestiondown Dónde se encuentra la capital de Reino de los Países Bajos? La respuesta es:'' (``What is the capital of the Kingdom of Netherlands? The answer is:'').}
        \label{fig:example-spanish}
        \includegraphics[width=0.9\linewidth]{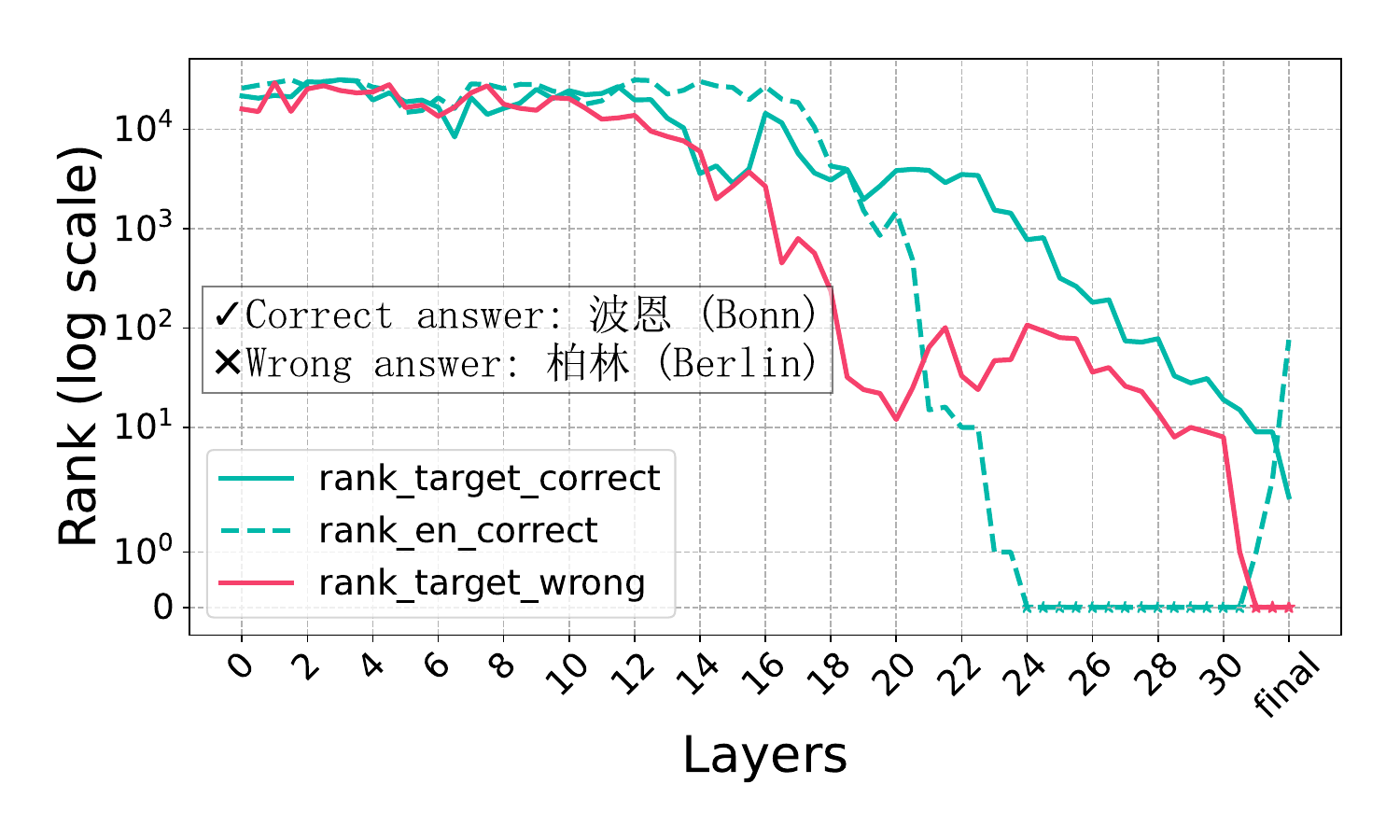}\caption{Prompt in Chinese: ``\begin{CJK}{UTF8}{gbsn}西德的首都在哪里？答案是：\end{CJK}'' (``What was the capital of West Germany? The answer is:'').}
        \label{fig:example-chinese}
    \end{subfigure}

    \caption{Rank evolution for prompts in Spanish (\ref{fig:example-spanish}) and Chinese (\ref{fig:example-chinese}). \texttt{rank\_target\_wrong} represents the rank of the model's final incorrect prediction across layers, while \texttt{rank\_target\_correct} and \texttt{rank\_en\_correct} denote the ranks of the correct answer in the target language and the English equivalent, respectively. The plots show the impact of errors during language transition, where the rank of the incorrect answer surpasses the correct answer in the final layers.
    }
    \label{fig:rank-plot-examples}
\end{figure}

\section{Examining the Cause of Cross-Lingual Inconsistency}
\label{sec:examine-inconsistency}

Next, we analyze incorrect predictions across languages to investigate the causes of cross-lingual inconsistencies in MLMs.

Figure~\ref{fig:mechanism-incorrect} shows the rank evolution for incorrect predictions in LLaMA2 and BLOOM. While the rank of the correct answer decreases significantly in the middle layers (both in the target language and in English) --- consistent with the behavior observed in correct predictions (Figure~\ref{fig:mechanism-row1}) --- the rank of the incorrect answer surpasses that of the correct answer during language transition in the final layers. This suggests that factual knowledge is processed in the concept space in the middle layers as in correct predictions, but errors arise during the transition to language-specific processing. 

To further investigate this phenomenon, we examine individual examples of LLaMA2.\footnote{LLaMA2's English-biased latent space provides clearer insights into the switch from English to the target language, while BLOOM's latent space is less interpretable, as shown in Figure~\ref{fig:language-composition-zh}.} Figure~\ref{fig:rank-plot-examples} presents cases in Spanish and Chinese, with additional examples provided in Appendix~\ref{sec:appendix-results-rank-predictions}. A consistent pattern emerges: in the middle-to-upper layers, the correct answer in English often ranks lowest (\texttt{rank\_en\_correct}=0), indicating accurate recall during the concept space processing stage. However, in the final layers, the rank of the incorrect target-language answer decreases, surpassing the correct answer during language transition. 

This observation underscores the critical role of language transition in cross-lingual inconsistencies. Although MLMs encode correct factual knowledge in the middle-layer concept space, the transition to language-specific processing introduces errors, causing incorrect predictions. Addressing this issue is crucial for improving cross-lingual consistency and robustness of MLMs.

\begin{figure}[h]
    \centering
    \includegraphics[width=1\linewidth]{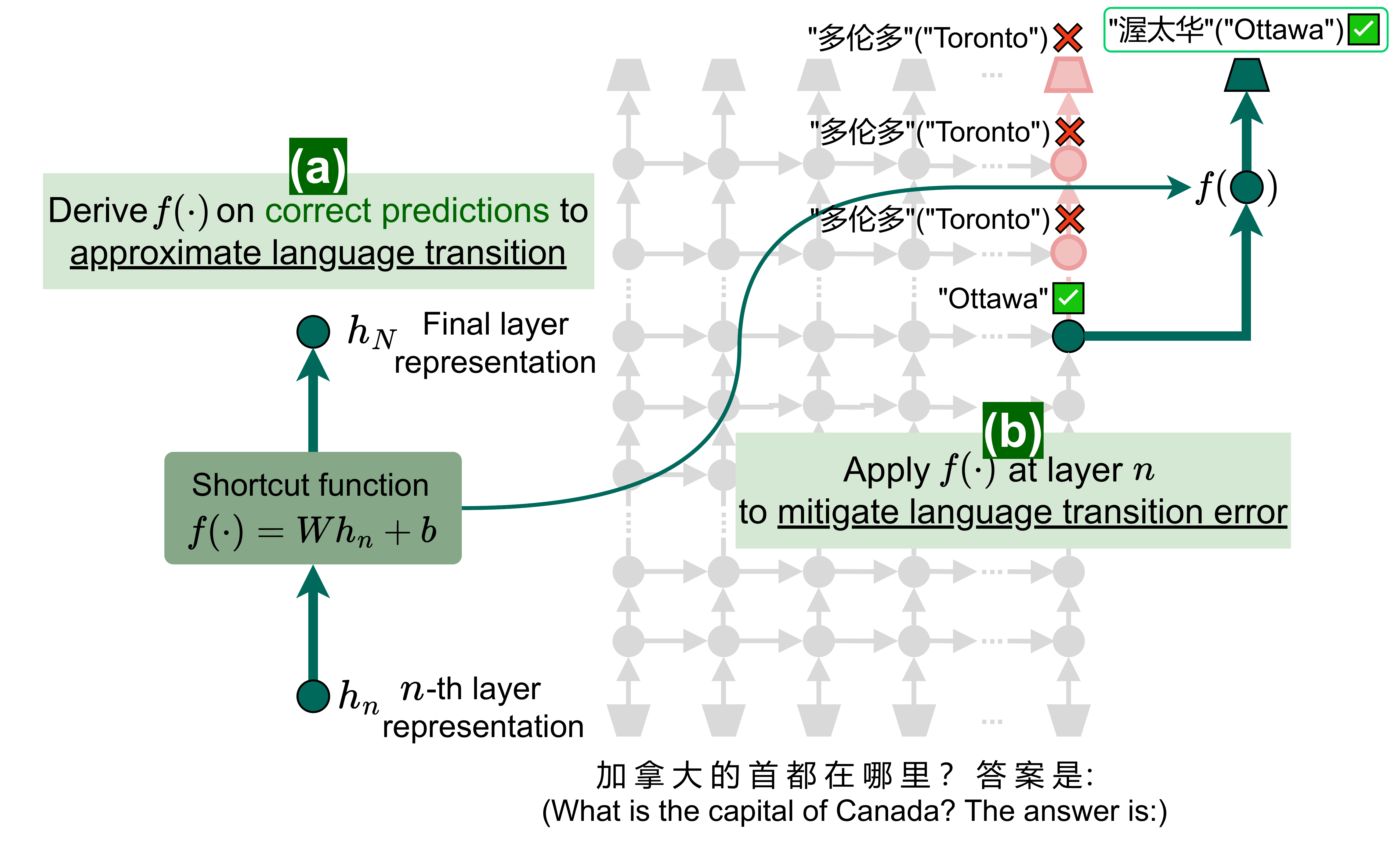}
    \caption{Illustration of the proposed shortcut method for mitigating cross-lingual inconsistency. \textbf{(a)} The shortcut function is learned on correct predictions to approximate language transition; \textbf{(b)} The learned function is then applied to bypass the error-prone final layers. In the example, the shortcut successfully recovers the correct answer, ``\begin{CJK}{UTF8}{gbsn}渥太华\end{CJK}'' (``Ottawa''), in Chinese.}
    \label{fig:shortcut-method}
\end{figure}

\section{Linear Shortcut for Improving Cross-Lingual Consistency}
\label{sec:shortcut}

In this section, we propose a linear shortcut method to address language transition errors. Our approach bypasses final-layer computations, directly adapting concept-space representations to the target language, enhancing both prediction accuracy and cross-lingual consistency of MLMs.

\subsection{
Shortcut with Linear Approximation}
\label{subsec:shortcut-method}


The proposed method consists two-step (illustrated in Figure~\ref{fig:shortcut-method}): (a) \textbf{Deriving the linear shortcut}: Inspired by \citet{hernandezlinearity}, we hypothesize that the mapping from the model's latent state at layer $n$ to the final layer $N$, i.e., $h_n \rightarrow h_N$ can be well-approximated by a linear function $f(h_n) = Wh_n + b \approx h_N$. Using $m$ correctly predicted samples per relation, we estimate $W$ and $b$ via first-order approximation, modeling how concept-space representations are adapted to the target language.\footnote{Layer $n$ and training size $m$ are treated as hyperparameters: $n=30$ for LLaMA2, $n=21$ for BLOOM, and $m=25$ for both models. Details are provided in Appendix~\ref{sec:appendix-shortcut-hp}.} We optimize one linear shortcut per language, shared across all relations, which aims to capture generalizable patterns in the representation-to-output transition for each language. Further details on the derivation and hyperparameters are provided in Appendix~\ref{sec:appendix-shortcut}. (b) \textbf{Applying the linear shortcut}: At inference time, the learned shortcut $f(\cdot)$ is applied to bypass the original final-layer computations, mitigating errors introduced during language transition.

\subsection{Results and Discussion}
\label{subsec:shortcut-res}
\begin{figure}
    \centering
    \includegraphics[width=0.95\linewidth]{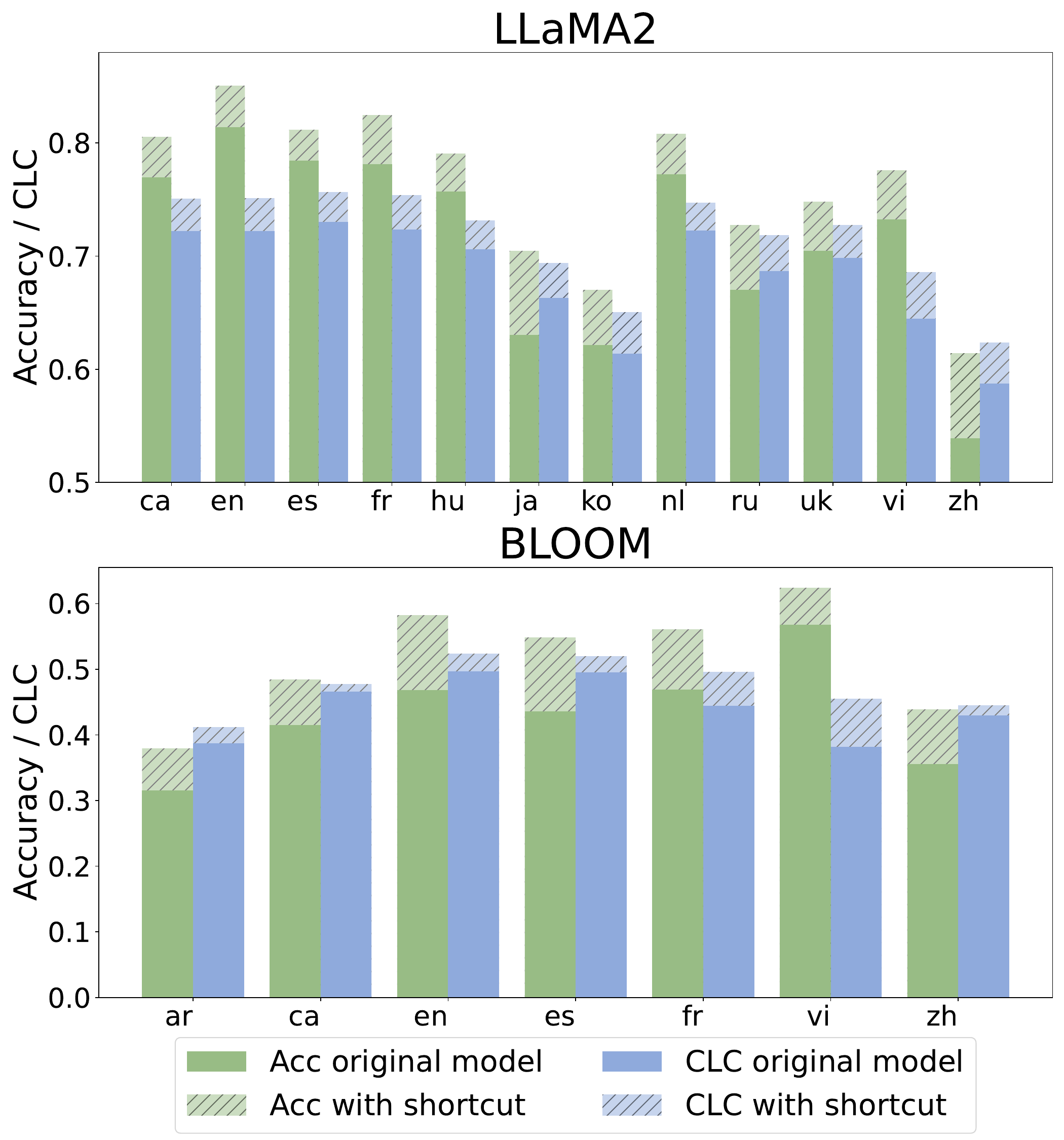}
    \caption{Accuracy (ACC) and cross-lingual consistency (CLC) per language for LLaMA2 and BLOOM, with and without the shortcut method.}
    \label{fig:shortcut-results}
\end{figure}

We evaluate the prediction accuracy and cross-lingual
consistency of LLaMA2 and BLOOM, without and with applying
the shortcut method, on all \klar samples. 

\paragraph{Baselines.}
We compare our shortcut method to three translation-based baselines: (1) \textit{translation-en}: We translate all input queries from each language to English using Google Translate, obtain model predictions in English, and then translate them back to the target language. (2) \textit{translation-early-exit}: We use Logit Lens to extract top-predicted tokens from the same layers as the shortcut method, translate them into the target language and evaluate their accuracy. (3) \textit{fine-tuning}: We fine-tune the models using $m = 25$ parallel samples per relation per language and evaluate on the full KLAR dataset, consistent with the settings used for our shortcut method. For efficiency, we applied LoRA-based fine-tuning to LLaMA2 (learning rate $lr = 1\text{e}{-4}$), and full model fine-tuning to BLOOM (learning rate $lr = 5\text{e}{-8}$) due to poor LoRA performance. Both models are trained with a batch size of 4 for 20 epochs.

\paragraph{Results.}
Figure~\ref{fig:shortcut-results} shows the effectiveness of the shortcut mapping: It improves prediction accuracy and cross-lingual consistency across models and languages. This demonstrates its ability to adapt concept-space knowledge to target languages for more reliable predictions.

\begin{table}[h!]
\footnotesize
\setlength{\tabcolsep}{3pt}
\centering
\begin{tabular}{l|ccccc} \toprule
      & \textbf{original} & \textbf{shortcut} & \textbf{trans-en} & \textbf{trans-exit} & \textbf{ft} \\ \midrule
\textbf{LLaMA2} & 71.47    & \textbf{76.08}    & 53.88    & 13.93  & 72.26    \\
\textbf{BLOOM} & 43.24    & \textbf{51.67}    & 28.03    & 15.68   & 31.73  \\ \bottomrule
\end{tabular}
\caption{Average accuracy across languages..}
\label{tab:shortcut-results-avg}
\end{table}

As shown in Table~\ref{tab:shortcut-results-avg}, both translation-based baseline methods perform poorly (see Table~\ref{tab:translation-baseline-llama2} and \ref{tab:translation-baseline-bloom} in Appendix for more details), indicating that existing translators are insufficient for cross-lingual factual prediction. 

Fine-tuning also yields unsatisfactory results. For LLaMA2, it slightly outperforms the original model but underperforms our shortcut method in accuracy. For BLOOM, fine-tuning underperforms the original model, with improvements seen only in English. We hypothesize that fine-tuning on a small subset of factual knowledge does not generalize well to unseen facts and may even degrade performance due to overfitting.

In contrast, our shortcut method directly adapts latent representations from earlier layers, preserving richer contextual information and thus improving prediction accuracy. Moreover, it is lightweight and efficient, relying only on linear operations, making it easily adaptable to existing MLMs.



\section{Conclusion}
This study investigates cross-lingual factual inconsistency in multilingual language models, revealing a three-stage knowledge recall process: language-independent relation processing, object extraction, and a final transition to language-specific adaptation. Errors in this transition often lead to incorrect predictions despite accurate object extraction. To address this, we propose a shortcut method that bypasses final-layer computations, improving prediction accuracy and cross-lingual consistency. Our findings enhance understanding of multilingual knowledge processing and introduce an efficient, interpretable solution for mitigating language transition errors.


Future work could expand the investigation to more languages and additional language models to assess broader applicability. Additionally, developing non-linear shortcut methods could better mitigate language transition errors, offering more robust solutions for cross-lingual consistency.

\section*{Limitations}
First, our cross-lingual consistency analysis assumes English as the pivot language, reflecting the English-centric nature of most multilingual models. While this aligns with prior studies \citep{wendler-etal-2024-llamas, dumas2024llamas, fierro2024multilingual}, it may limit applicability to language pairs that do not involve English. 

Second, although the KLAR dataset covers 17 languages, it does not fully capture the diversity of world languages. Expanding the analysis to more languages and exploring models with different architectures and sizes could provide deeper insights into cross-lingual inconsistencies. 

Additionally, our shortcut method relies on linear approximation for simplicity. Investigating non-linear approaches could better capture complex transformations during language switching and further enhance performance. 

Finally, our analysis provides insights relevant to downstream tasks, such as multilingual knowledge localization \citep{chen2024journey, kojima-etal-2024-multilingual, tang-etal-2024-language} and cross-lingual knowledge editing \citep{xu-etal-2023-language-anisotropic, nie2024bmike}. 
However, these applications fall beyond the scope of this study and are left for future work.

\section*{Acknowledgments}
This work was partially supported by Deutsche Forschungsgemeinschaft (project SCHU 2246/14-1). We would like to thank Sebastian Gerstner, Ahmad Dawar Hakimi, Anna Hätty, Lea Hirlimann, Amir Hossein Kargaran, Valentin Knappich, Felicia Körner, Mohsen Mesgar, Ali Modarressi, Philipp Mondorf, Timo Pierre Schrader, Leonor Veloso, Wei Zhou, Yuqicheng Zhu for fruitful discussions.

\section*{Ethical considerations}
This study investigates cross-lingual factual inconsistencies in multilingual language models. While our focus is diagnostic, incorrect model predictions may propagate misinformation or reflect underlying biases present in the models.

\bibliography{anthology,custom}

\newpage

\appendix

\section{Appendix}
\label{sec:appendix}

\begin{table*}[htbp]
  \footnotesize
  \centering
    \begin{tabular}{l|c|l} \toprule
    \textbf{Relation} & \textbf{\# Facts} & \textbf{Prompt Example} \\
    \midrule
    applies\_to\_jurisdiction & 79    & Which country has <subject> as a legal term? The answer is: \\
    capital & 336   & What is the capital of <subject>? The answer is: \\
    capital\_of & 212   & Where is <subject> the capital of? The answer is: \\
    continent & 212   & Which continent is <subject> located in? The answer is: \\
    country\_of\_citizenship & 60    & Which country is <subject> a citizen of? The answer is: \\
    developer & 76    & Which company is the developer of <subject>? The answer is: \\
    field\_of\_work & 167   & What field does <subject> work in? The answer is: \\
    headquarters\_location & 51    & In which city is <subject>'s headquarter located? The answer is: \\
    instrument & 46    & Which musical instrument is played by <subject>? The answer is: \\
    language\_of\_work\_or\_name & 108   & What is the original language of <subject>? The answer is: \\
    languages\_spoken & 104   & What language did <subject> use to communicate? The answer is: \\
    location\_of\_formation & 66    & Where did the formation of <subject> take place? The answer is: \\
    manufacturer & 35    & Which company manufactures <subject>? The answer is: \\
    native\_language & 130   & What is the native language of <subject>? The answer is: \\
    occupation & 46    & What is <subject>'s profession? The answer is: \\
    official\_language & 602   & What is the official language of <subject>? The answer is: \\
    owned\_by & 50    & Who is the current owner of <subject>? The answer is: \\
    place\_of\_birth & 35    & In which city was <subject> born? The answer is: \\
    place\_of\_death & 79    & In which city did <subject> pass away? The answer is: \\
    religion & 125   & What is the religious belief of <subject>? The answer is: \\
    \bottomrule
    \end{tabular}%
  \caption{Relations in the \klar dataset with fact counts and prompt examples used for knowledge probing.}
  \label{tab:addlabel}%
\end{table*}

\subsection{\klar Dataset Details}
\label{sec:appendix-dataset}

As discussed in Section~\ref{sec:dataset}, BMLAMA17 \citep{qi-etal-2023-cross} is incompatible with multilingual knowledge probing in auto-regressive models with many objects placed in the middle of sentences, and many relations types with multiple correct answers. To address these limitations, we construct \klar for reliable multilingual knowledge probing evaluation.

BMLAMA17 does not explicitly specify relation types; however, many factual questions share the same templates. We first group sentences with identical templates and use \texttt{gpt-3.5-turbo} to identify the relation for each template and map them to Wikidata property IDs \citep{wikidata_properties}. We discard the samples which cannot be mapped to any Wikidata property. This process yields a total of 42 relation types. 

For each relation, we generate English prompt templates in the format of ``<\textit{Question}> The answer is:'' as introduced in Section~\ref{sec:dataset}, using \texttt{gpt-3.5-turbo}. We created five templates per relation and manually verify their clarity.
The templates are then translated into 16 additional languages using \texttt{gpt-3.5-turbo}. Their quality is manually reviewed for Chinese, Spanish, and Japanese. Back-translation is used to verify clarity and consistency in the remaining languages.

Finally, we remove relation types with multiple correct answers and those with fewer than 30 samples. The resulting \klar dataset comprises parallel factual knowledge spanning 17 languages and 20 relation types. For the analysis on LLaMA2 and BLOOM models, we use the intersection of languages supported by these models and included in \klar, covering 12 languages for LLaMA2 and 7 for BLOOM, see Table~\ref{tab:klar-languages-all} for the respective language list. Listing~\ref{lst:dataset-example} illustrates the example of the \klar dataset structure for the relation \textit{capital} in English.

\begin{table*}[h]
  \footnotesize
  \centering
    \begin{tabular}{lp{10.5cm}} \toprule
    \textbf{\klar languages (17)} & Arabic (ar), Catalan(ca), Greek (el), English (en), Spanish (es), Persian (fa), French (fr), Hebrew (he), Hungarian (hu), Japanese (ja), Korean (ko), Dutch (nl), Russian (ru), Turkish (tr), Ukrainian (uk), Vietnamese (vi), Chinese (zh) \\
    \midrule
    \textbf{LLaMA2 overlap (12)} & Catalan(ca), English (en), Spanish (es), French (fr), Hungarian (hu), Japanese (ja), Korean (ko), Dutch (nl), Russian (ru), Ukrainian (uk), Vietnamese (vi), Chinese (zh) \\
    \midrule
    \textbf{BLOOM overlap (7)} & Arabic (ar), Catalan(ca), English (en), Spanish (es), French (fr), Vietnamese (vi), Chinese (zh) \\
    \bottomrule
    \end{tabular}%
  \caption{\klar dataset languages and their overlap with LLaMA2 and BLOOM.}
  \label{tab:klar-languages-all}%
\end{table*}%

\begin{lstlisting}[language=json, caption={Example of \klar for relation \textit{capital} in English.}, label={lst:dataset-example}, captionpos=b]
{
    "relation_name": "capital",
    "relation_id": "P36",
    "prompt_templates": [
        "Where is <subject>'s capital located? The answer is:",
        "What is the capital of <subject>? The answer is:",
        "Which city serves as the capital of <subject>? The answer is:",
        "Name the capital city of <subject>. The answer is:",
        "Where does <subject> have its capital? The answer is:"
    ],
    "samples": [
        {
            "subject": "Azerbaijan",
            "object": "Baku",
            "index": 6152
        },
        {
            "subject": "Germany",
            "object": "Berlin",
            "index": 6165
        },
    ]
}
\end{lstlisting}

\subsection{Additional Experimental Results}
\label{sec:appendix-results}

\subsubsection{Latent State Similarity}
\label{sec:appendix-similarity}

\begin{figure}[h]
    \centering
    \begin{subfigure}{\linewidth}
        \centering
        \includegraphics[width=0.96\linewidth]{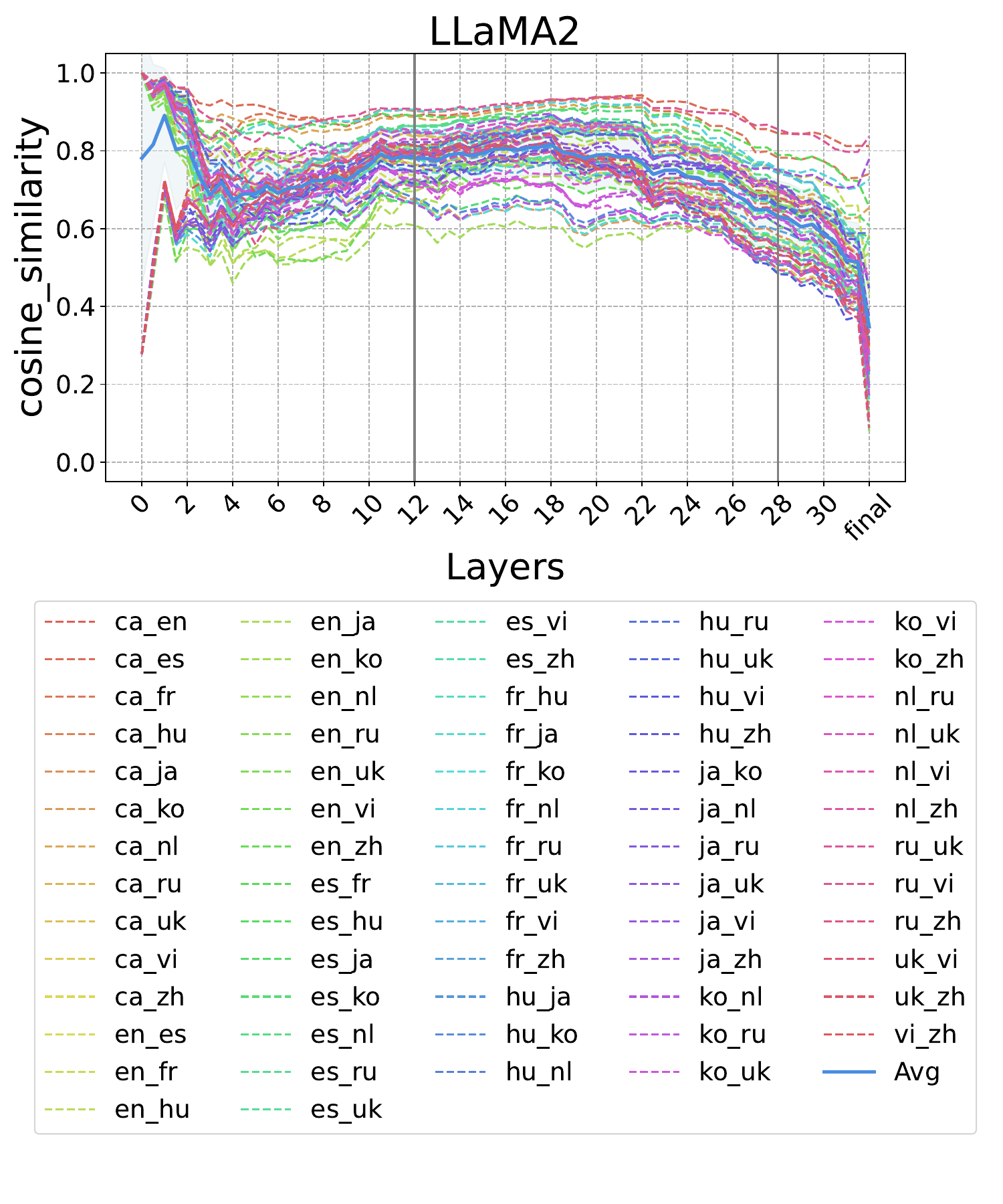}
        \includegraphics[width=0.96\linewidth]{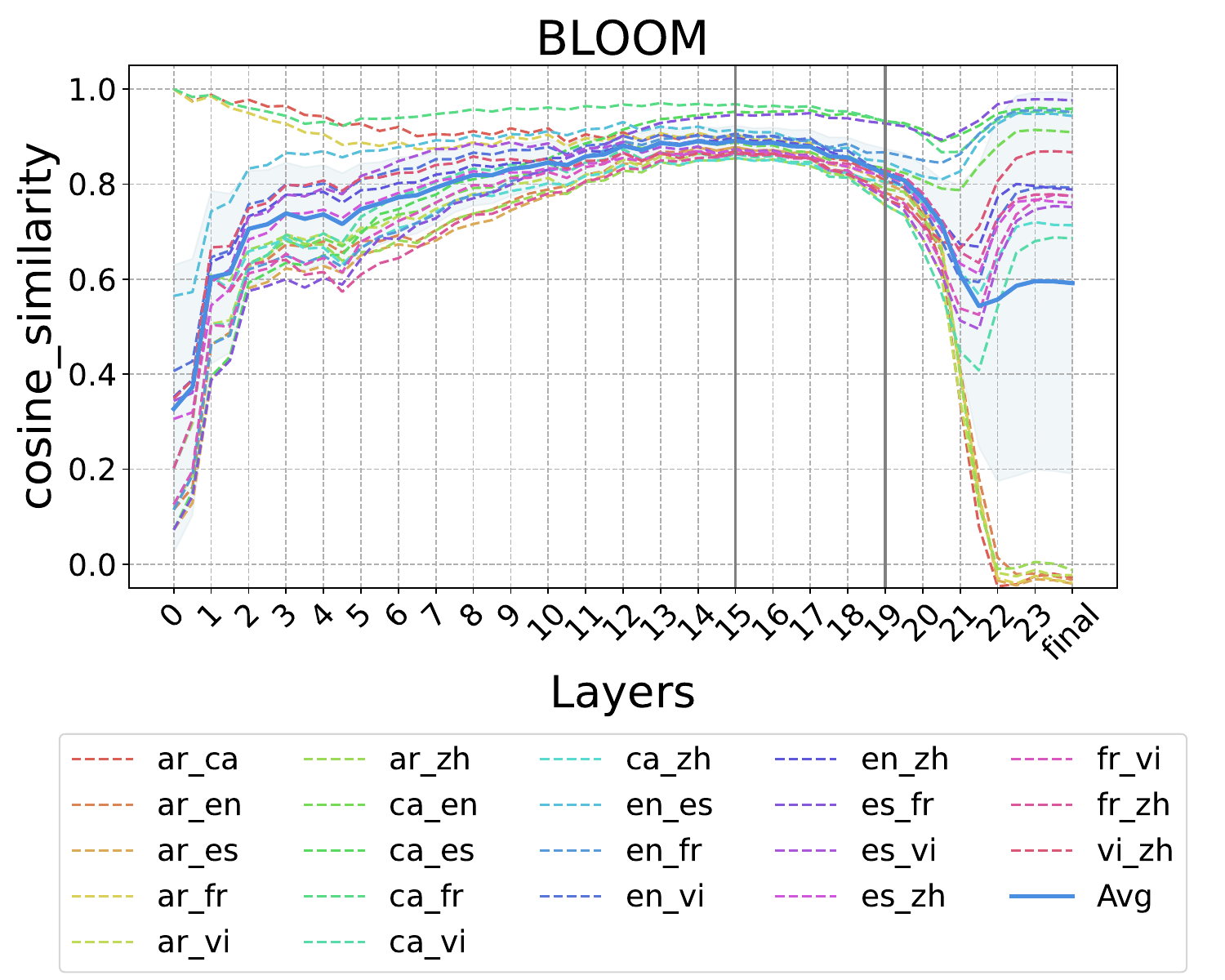}
    \end{subfigure}

    \caption{Cosine similarity of latent states between all language pairs averaged across all relation.}
    \label{fig:latent-similarity-all-pairs}
\end{figure}

Here, we present the complete results for latent state similarity across all language pairs in Figure~\ref{fig:latent-similarity-all-pairs}.

The plots follow the same trend as in Figure~\ref{fig:mechanism-row2}, where similarity across language pairs increases from early to middle layers in both models, indicating that MLMs encode information in a concept space independent of the input language. In the final layers, similarity declines as representations transition to a language-specific form. This pattern holds even for linguistically diverse pairs, highlighting that MLMs initially process factual knowledge in a shared latent space before adapting it to the target language.

\subsubsection{Latent Space Language Composition}
\label{sec:appendix-results-composition}
We examine the language composition of the latent states in LLaMA2 and BLOOM to understand how these MLMs encode information in the concept space. As described in Section~\ref{subsec:language-composition}, we apply Logit Lens to project latent states to the vocabulary, and use fasttext to identify the language of the top-10 predicted tokens at each layer.

Figure~\ref{fig:language-composition-shared} presents results for languages shared between LLaMA2 and BLOOM, while Figure~\ref{fig:language-composition-unique} shows results for languages unique to each model.

LLaMA2's middle-to-upper layers are dominated by English, aligning with prior findings that ``LLaMA2 models think in English'' \citep{wendler-etal-2024-llamas}. In contrast, BLOOM displays a more diverse linguistic composition in these layers.

Across different input languages, both models exhibit similar language distributions in the middle-to-upper layers, indicating that MLMs encode knowledge in a concept space largely independent of the input language.

\subsubsection{Rank Plots of Wrong Predictions}
\label{sec:appendix-results-rank-predictions}
Figure~\ref{fig:app-rank-plot-examples} presents additional examples, one per language, where the correct English answer ranks highest in the middle-to-upper layers but is later surpassed by an incorrect target-language answer during the language transition phase.

\subsection{Shortcut Experimental Details}
\label{sec:appendix-shortcut}
\subsubsection{Method} 
The idea of using linear approximation as a shortcut is inspired by \citet{hernandezlinearity}, who derive a linear transformation to approximate the mapping from subject to object representations in factual knowledge, showing that relational decoding in transformer models can be effectively modeled with linear functions.

Building on this idea, we apply linear approximation to address cross-lingual inconsistency by bypassing the language transition process in MLMs. We hypothesize that the mapping from the model's latent state at layer $n$ to that at the final layer $N$, i.e., $h_n \rightarrow h_N$ can be well-approximated by a linear function $f(h_n) = Wh_n + b \approx h_N$. Following \citet{hernandezlinearity}, we use first-order approximation to estimate $W_r$ and $b_r$ as the mean Jacobian and bias across $m$ correctly predicted factual samples $\{h_{n_i}, h_{N_i}\}_{i=1, \dots, m}$. That is, we define: 

\begin{equation}
\begin{split}
    W_r &= \mathbb{E}_{h_{n_i}, h_{N_i}} \left[ \frac{\partial F}{\partial h_n} \bigg|_{(h_{n_i}, h_{N_i})} \right], \\
    b_r &= \mathbb{E}_{h_{n_i}, h_{N_i}} \left[ h_N - \frac{\partial F}{\partial h_n} \bigg|_{(h_{n_i}, h_{N_i})} h_n \right]
\end{split}
\end{equation}

As noted in \citet{hernandezlinearity}, the first-order derivative $W_r$ tends to underestimate the magnitude of changes from $h_n$ to $h_N$ in practice. They attribute this to the use of layer normalization \citep{lei2016layer} in transformers: which does not transmit changes in scale of inputs to changes in scale of output. Specifically, the input $h_n$ at layer $n$ is normalized before being propagated to subsequent layers. To address this underestimation, a scalar constant $\beta$ is introduced as a hyperparameter and multiplied by $W_r$ as a corrective factor:

\begin{equation}
f(h_n) = \beta W_rh_n + b_r = Wh_n + b
\end{equation}

\subsubsection{Hyperparameters}
\label{sec:appendix-shortcut-hp}
Several hyperparameters are introduced when determining the linear shortcut $f(\cdot)$: the layer $n$ from which the latent state is extracted for linear approximation, the scalar constant $\beta$ used to adjust the slope of $W_r$ to account for the underestimation in the first-order approximation of $h_n \rightarrow h_N$, and the number of correct samples used to compute $f(\cdot)$.
We perform a grid search to select these hyperparameters per language, aiming to maximize prediction accuracy. For the layer $n$, we search within the range of $[20, 32]$ for LLaMA2 and $[12, 24]$ for BLOOM. The scalar constant $\beta$ is searched over the range $[0, 5.0]$ in increments of 0.25, following \citet{hernandezlinearity}. The number of samples $m$ is selected from $[10, 25, 40, 50]$. 
The hyperparameter search is conducted for each language individually. We find that the optimal $\beta$ value varies across languages, while the other two hyperparameters --- the extraction layer $n$ and the number of samples $m$ --- remain consistent across languages. The selected hyperparameters for both models are summarized in Table~\ref{tab:shortcut-hp-llama2} and \ref{tab:shortcut-hp-bloom}, respectively.

\begin{table}[h]
    \footnotesize
    \centering
    \scalebox{0.9}{
    \begin{tabular}{c|>{\centering\arraybackslash}p{1.25cm}>{\centering\arraybackslash}p{1.25cm}>{\centering\arraybackslash}p{1.25cm}} 
        \toprule
        \textbf{LLaMA2} & \textbf{$n$} & \textbf{$\beta$} & \textbf{$m$} \\ 
        \midrule
        \textbf{ca} & \multirow{12}{*}{30} & 4.75 & \multirow{12}{*}{25} \\
        \textbf{en} & & 1.50 & \\
        \textbf{es} & & 3.00 & \\
        \textbf{fr} & & 4.25 & \\
        \textbf{hu} & & 2.50 & \\
        \textbf{ja} & & 2.25 & \\
        \textbf{ko} & & 4.50 & \\
        \textbf{nl} & & 3.50 & \\
        \textbf{ru} & & 4.25 & \\
        \textbf{uk} & & 2.25 & \\
        \textbf{vi} & & 1.00 & \\
        \textbf{zh} & & 1.50 & \\
        \bottomrule
    \end{tabular}
    }
    \caption{Hyperparameters per language for LLaMA2.}
    \label{tab:shortcut-hp-llama2}
\end{table}

\begin{table}[h]
    \footnotesize
    \centering
    \scalebox{0.9}{
    \begin{tabular}{c|>{\centering\arraybackslash}p{1.25cm}>{\centering\arraybackslash}p{1.25cm}>{\centering\arraybackslash}p{1.25cm}} 
        \toprule
        \textbf{BLOOM} & \textbf{$n$} & \textbf{$\beta$} & \textbf{$m$} \\ 
        \midrule
        \textbf{ar} & \multirow{7}{*}{21} & 1.25 & \multirow{7}{*}{25} \\
        \textbf{ca} & & 1.00 & \\
        \textbf{en} & & 1.25 & \\
        \textbf{es} & & 1.00 & \\
        \textbf{fr} & & 0.75 & \\
        \textbf{vi} & & 1.25 & \\
        \textbf{zh} & & 1.50 & \\
        \bottomrule
    \end{tabular}
    }
    \caption{Hyperparameters per language for BLOOM.}
    \label{tab:shortcut-hp-bloom}
\end{table}

\begin{table*}[!htbp]
\centering
\footnotesize
\begin{tabular}{l|ccc|ccc}
\toprule
\multirow{2}{*}{\textbf{Relations}} &
\multicolumn{3}{c|}{\textbf{Accuracy (acc)}} &
\multicolumn{3}{c}{\textbf{Cross-lingual Consistency (clc)}} \\
& \textbf{Original} & \textbf{Shortcut} & \textbf{Diff}
& \textbf{Original} & \textbf{Shortcut} & \textbf{Diff} \\
\midrule
applies\_to\_jurisdiction & 92.92 & \textbf{96.28} & 3.36 & 87.60 & \textbf{92.84} & 5.24 \\
capital & 83.06 & \textbf{88.54} & 5.48 & 80.87 & \textbf{86.10} & 5.23 \\
capital\_of & 66.16 & \textbf{70.26} & 4.10 & 71.66 & \textbf{74.98} & 3.32 \\
continent & 85.50 & \textbf{90.37} & 4.87 & 81.34 & \textbf{86.94} & 5.60 \\
country\_of\_citizenship & 71.53 & \textbf{76.38} & 4.85 & 69.41 & \textbf{72.91} & 3.50 \\
developer & 90.90 & \textbf{94.05} & 3.15 & 84.02 & \textbf{87.12} & 3.10 \\
field\_of\_work & 47.50 & \textbf{53.39} & 5.89 & 46.34 & \textbf{53.99} & 7.65 \\
headquarters\_location & 68.79 & \textbf{74.40} & 5.61 & 62.94 & \textbf{67.28} & 4.34 \\
instrument & 60.87 & \textbf{65.22} & 4.35 & 66.48 & \textbf{72.76} & 6.28 \\
language\_of\_work\_or\_name & 84.49 & \textbf{88.09} & 3.60 & 86.14 & \textbf{90.68} & 4.54 \\
languages\_spoken & 75.48 & \textbf{83.65} & 8.17 & 71.89 & \textbf{81.64} & 9.75 \\
location\_of\_formation & 49.24 & \textbf{56.56} & 7.32 & 44.81 & \textbf{49.58} & 4.77 \\
manufacturer & 94.28 & \textbf{96.83} & 2.55 & 91.77 & \textbf{93.97} & 2.18 \\
native\_language & 91.09 & \textbf{93.24} & 2.15 & 87.75 & \textbf{92.50} & 4.75 \\
occupation & 36.23 & \textbf{42.22} & 5.99 & 48.18 & \textbf{56.50} & 8.32 \\
official\_language & 67.64 & \textbf{71.22} & 3.58 & 74.14 & \textbf{77.32} & 3.18 \\
owned\_by & 60.50 & \textbf{64.59} & 4.09 & 57.27 & \textbf{62.57} & 5.30 \\
place\_of\_birth & 53.33 & \textbf{57.95} & 4.62 & 47.89 & \textbf{54.26} & 6.37 \\
place\_of\_death & 67.62 & \textbf{72.89} & 5.27 & 66.15 & \textbf{69.95} & 3.80 \\
religion & 82.23 & \textbf{85.33} & 3.10 & 82.15 & \textbf{86.41} & 4.26 \\
\midrule
\textbf{AVG} & 71.47 & \textbf{76.08} & 4.60 & 70.44 & \textbf{75.52} & 5.07 \\
\bottomrule
\end{tabular}
\caption{Prediction accuracy (acc) and cross-lingual consistency (clc) of LLaMA2 before and after applying the shortcut method across different relations.}
\label{tab:llama2-per-relation-merged}
\end{table*}

\subsubsection{Shortcut Translation Baselines.}
\label{sec:appendix-translation-baseline}

As mentioned in Section~\ref{subsec:shortcut-res}, we compare our shortcut method with two translation-based baselines: (1) translation-en (trans-en): We translate all input queries from each language to English using Google Translate, obtain model predictions in English, and then translate them back to the target language to measure accuracy. (2) translation-early-exit (trans-exit): We use Logit Lens to project the latent states at the same extraction layers as in the shortcut method, i.e., layer 30 for LLaMA2 and layer 20 for BLOOM, and extract the top-predicted tokens. These tokens are then translated into the target language using Google Translate, and their accuracy is calculated against the correct object.

As shown in Table~\ref{tab:translation-baseline-llama2} and \ref{tab:translation-baseline-bloom}, both translation-based methods perform poorly. The low accuracy of \textit{translation-en} suggests that existing translators struggle with entity translation, especially for languages that are highly dissimilar to English. The poor performance of \textit{translation-early-exit} stems from the inherent unreliability of token-level translations. Overall, these results indicate that translation-based approaches are not a viable solution for cross-lingual factual prediction.  
In contrast, by directly adapting latent representations from earlier layers, the shortcut method operates at the representation level, capturing richer contextual information. This enables significantly higher prediction accuracy and offers a promising solution for mitigating cross-lingual factual inconsistency.

\begin{table}[h]
  \footnotesize
  \centering
  \scalebox{0.8}{
    \begin{tabular}{c|ccccc}
    \toprule
    \textbf{LLaMA2} & \textbf{original} & \textbf{shortcut} & \textbf{trans-en} & \textbf{trans-exit} & \textbf{ft}\\
    \midrule
    \textbf{ca} & 76.96 & \textbf{80.54} & 44.95 & 24.52 & 70.25\\
    \textbf{en} & 81.41 & \textbf{85.06} & 81.41 & 43.05 & 80.43\\
    \textbf{es} & 78.44 & \textbf{81.16} & 47.77 & 28.42 & 75.44\\
    \textbf{fr} & 78.14 & \textbf{82.46} & 53.27 & 24.85 & 76.58\\
    \textbf{hu} & 75.69 & \textbf{79.04} & 64.60 & 6.91 & 70.48\\
    \textbf{ja} & 63.05 & \textbf{70.45} & 59.59 & 0.13 & 63.51\\
    \textbf{ko} & 62.14 & \textbf{66.98} & 49.30 & 0.28 & 52.38 \\
    \textbf{nl} & 77.22 & \textbf{80.77} & 62.07 & 15.24 & 75.1\\
    \textbf{ru} & 67.02 & \textbf{72.71} & 47.58 & 2.72 & 67.56 \\
    \textbf{uk} & 70.46 & \textbf{74.78} & 46.59 & 5.62 & 67.18\\
    \textbf{vi} & 73.26 & \textbf{77.56} & 39.07 & 12.70 & 70.82\\
    \textbf{zh} & 53.88 & \textbf{61.40} & 60.38 & 1.67 & 52.66\\
    \bottomrule
    \end{tabular}%
    }
  \caption{Comparison of the prediction accuracy (\%) for LLaMA2 across different languages using the original model, the proposed shortcut method, and the translation-based baselines.}
  \label{tab:translation-baseline-llama2}%
\end{table}%

\begin{table}[h]
  \footnotesize
  \centering
  \scalebox{0.8}{
    \begin{tabular}{c|ccccc}
    \toprule
    \textbf{BLOOM} & \textbf{original} & \textbf{shortcut} & \textbf{trans-en} & \textbf{trans-exit} & \textbf{ft}\\
    \midrule
    \textbf{ar} & 31.58 & \textbf{37.93} & 21.87 & 0.97 & 23.99\\
    \textbf{ca} & 41.50  & \textbf{48.40} & 22.58 & 15.88 & 28.93\\
    \textbf{en} & 46.81 & \textbf{58.24} & 46.81 & 26.85 & 49.97\\
    \textbf{es} & 43.56 & \textbf{54.84} & 25.53 & 11.26 & 36.17\\
    \textbf{fr} & 46.88 & \textbf{56.03} & 26.15 & 17.97 & 34.14 \\
    \textbf{vi} & 56.82 & \textbf{62.38} & 21.98 & 25.85 & 29.54 \\
    \textbf{zh} & 35.54 & \textbf{43.89} & 31.26 & 10.96 & 25.44 \\
    \bottomrule
    \end{tabular}%
    }
  \caption{Comparison of the prediction accuracy (\%) for BLOOM across different languages using the original model, the proposed shortcut method, and the translation-based baselines.}
  \label{tab:translation-baseline-bloom}%
\end{table}%

\subsubsection{Per-relation Shortcut Performance.}
\label{sec:appendix-shortcut-per-relation}

In Tables~\ref{tab:llama2-per-relation-merged} and \ref{tab:bloom-per-relation-merged}, we provide a detailed per-relation breakdown of performance for both the original LLaMA2 and BLOOM models and their shortcut-enhanced counterparts, covering prediction accuracy (acc) and cross-lingual consistency (clc).

The results demonstrate that the improvements are not limited to a specific relation, but are consistently observed across a wide range of relation types, underscoring the robustness and generalizability of the proposed shortcut method.

\begin{table*}[!htbp]
\centering
\footnotesize
\begin{tabular}{l|ccc|ccc}
\toprule
\multirow{2}{*}{\textbf{Relations}} &
\multicolumn{3}{c|}{\textbf{Accuracy (acc)}} &
\multicolumn{3}{c}{\textbf{Cross-lingual Consistency (clc)}} \\
& \textbf{Original} & \textbf{Shortcut} & \textbf{Diff}
& \textbf{Original} & \textbf{Shortcut} & \textbf{Diff} \\
\midrule
applies\_to\_jurisdiction & 88.57 & \textbf{93.56} & 4.99 & 87.17 & \textbf{90.03} & 2.86 \\
capital & 42.86 & \textbf{48.60} & 5.74 & 47.44 & \textbf{52.67} & 5.23 \\
capital\_of & 36.28 & \textbf{42.35} & 6.07 & 40.77 & \textbf{45.96} & 5.19 \\
continent & 18.73 & \textbf{55.73} & 37.00 & 19.68 & \textbf{39.58} & 19.90 \\
country\_of\_citizenship & 32.38 & \textbf{43.81} & 11.43 & 36.33 & \textbf{44.86} & 8.53 \\
developer & 74.06 & \textbf{78.26} & 4.20 & 67.52 & \textbf{73.45} & 5.93 \\
field\_of\_work & 12.92 & \textbf{22.50} & 9.58 & 24.05 & \textbf{34.08} & 9.83 \\
headquarters\_location & 31.09 & \textbf{36.41} & 5.32 & 49.89 & \textbf{53.96} & 4.07 \\
instrument & 46.27 & \textbf{52.56} & 6.29 & 97.84 & \textbf{98.87} & 1.03 \\
language\_of\_work\_or\_name & 62.30 & \textbf{69.04} & 6.74 & 75.24 & \textbf{79.65} & 4.41 \\
languages\_spoken & 47.66 & \textbf{53.37} & 5.71 & 59.66 & \textbf{65.44} & 5.78 \\
location\_of\_formation & 17.32 & \textbf{22.29} & 4.97 & 27.55 & \textbf{32.41} & 4.86 \\
manufacturer & 88.16 & \textbf{92.61} & 4.45 & 83.07 & \textbf{91.02} & 7.95 \\
native\_language & 54.40 & \textbf{70.99} & 16.59 & 38.52 & \textbf{48.71} & 10.19 \\
occupation & 20.19 & \textbf{26.25} & 6.06 & 37.56 & \textbf{42.86} & 5.30 \\
official\_language & 47.01 & \textbf{54.14} & 7.13 & 44.03 & \textbf{49.18} & 5.15 \\
owned\_by & 40.00 & \textbf{45.71} & 5.71 & 52.11 & \textbf{57.50} & 5.39 \\
place\_of\_birth & 19.59 & \textbf{26.25} & 6.66 & 44.94 & \textbf{49.03} & 4.09 \\
place\_of\_death & 23.91 & \textbf{40.12} & 16.21 & 53.61 & \textbf{60.36} & 6.75 \\
religion & 52.11 & \textbf{58.12} & 6.01 & 82.93 & \textbf{84.62} & 1.69 \\
\midrule
\textbf{AVG} & 43.24 & \textbf{51.67} & 8.43 & 54.16 & \textbf{60.32} & 6.16 \\
\bottomrule
\end{tabular}
\caption{Prediction accuracy (acc) and cross-lingual consistency (clc) of BLOOM before and after applying the shortcut method across different relations.}
\label{tab:bloom-per-relation-merged}
\end{table*}

\begin{figure*}[h]
\begin{subfigure}{0.48\linewidth}
    \centering
    \includegraphics[width=\linewidth]{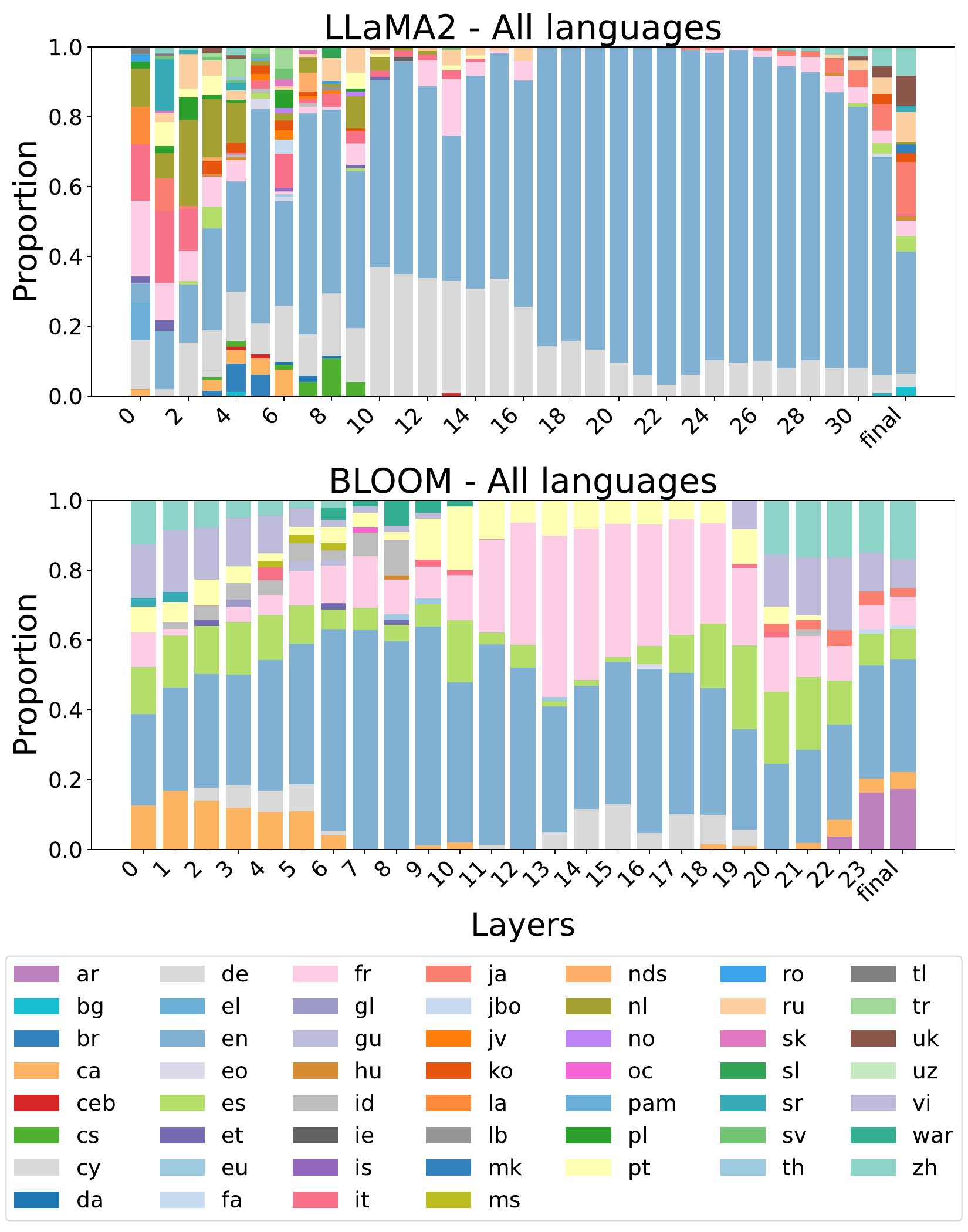}
    \caption{Language composition aggregated across all languages}
    \label{fig:language-composition-all}
\end{subfigure}
\hfill
\begin{subfigure}{0.48\linewidth}
    \centering
    \includegraphics[width=\linewidth]{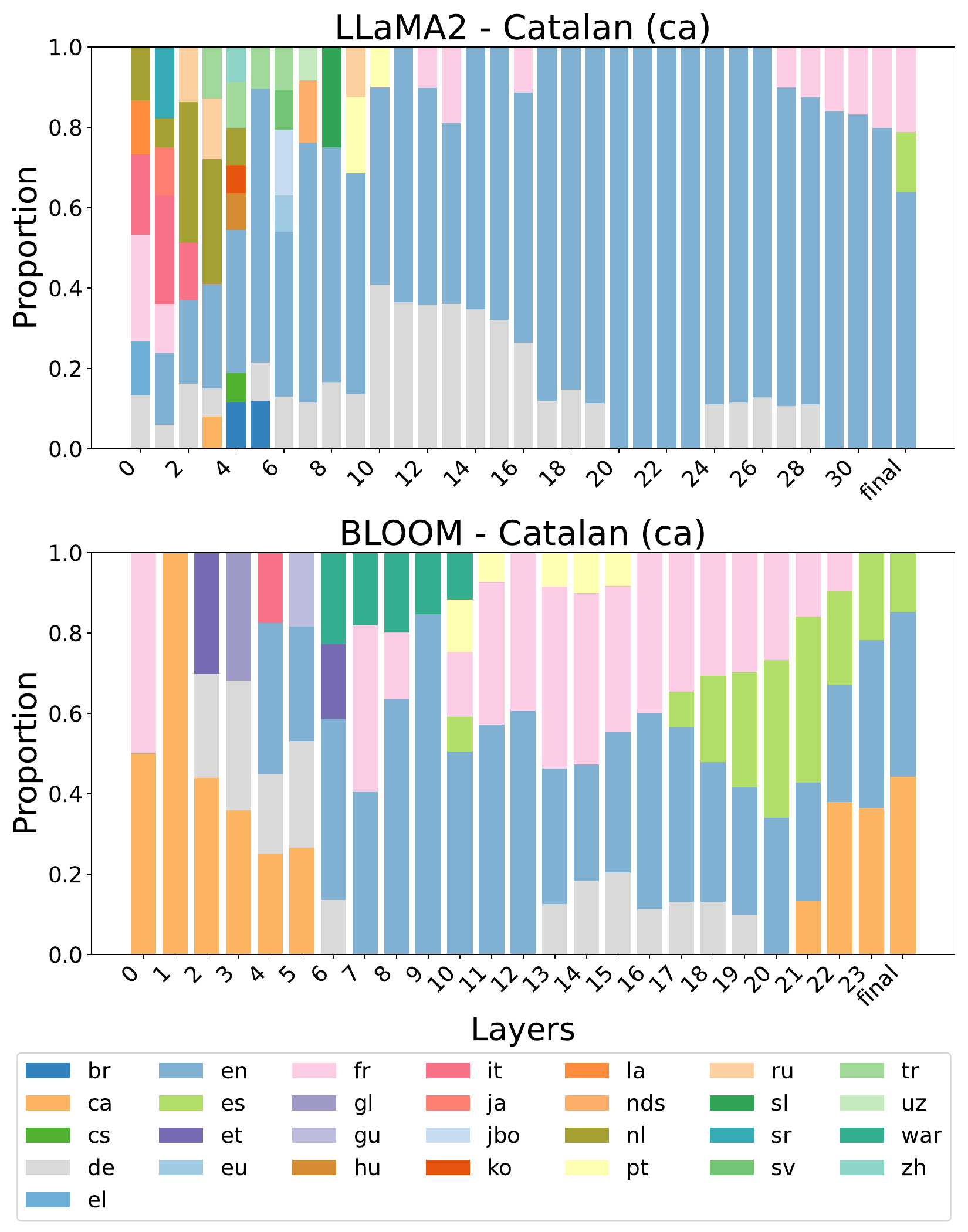}
    \caption{Language composition with Catalan as the input language.}
    \label{fig:language-composition-ca}
\end{subfigure}
\end{figure*}
\begin{figure*}[h]
\ContinuedFloat
\begin{subfigure}{0.48\linewidth}
    \centering
    \includegraphics[width=\linewidth]{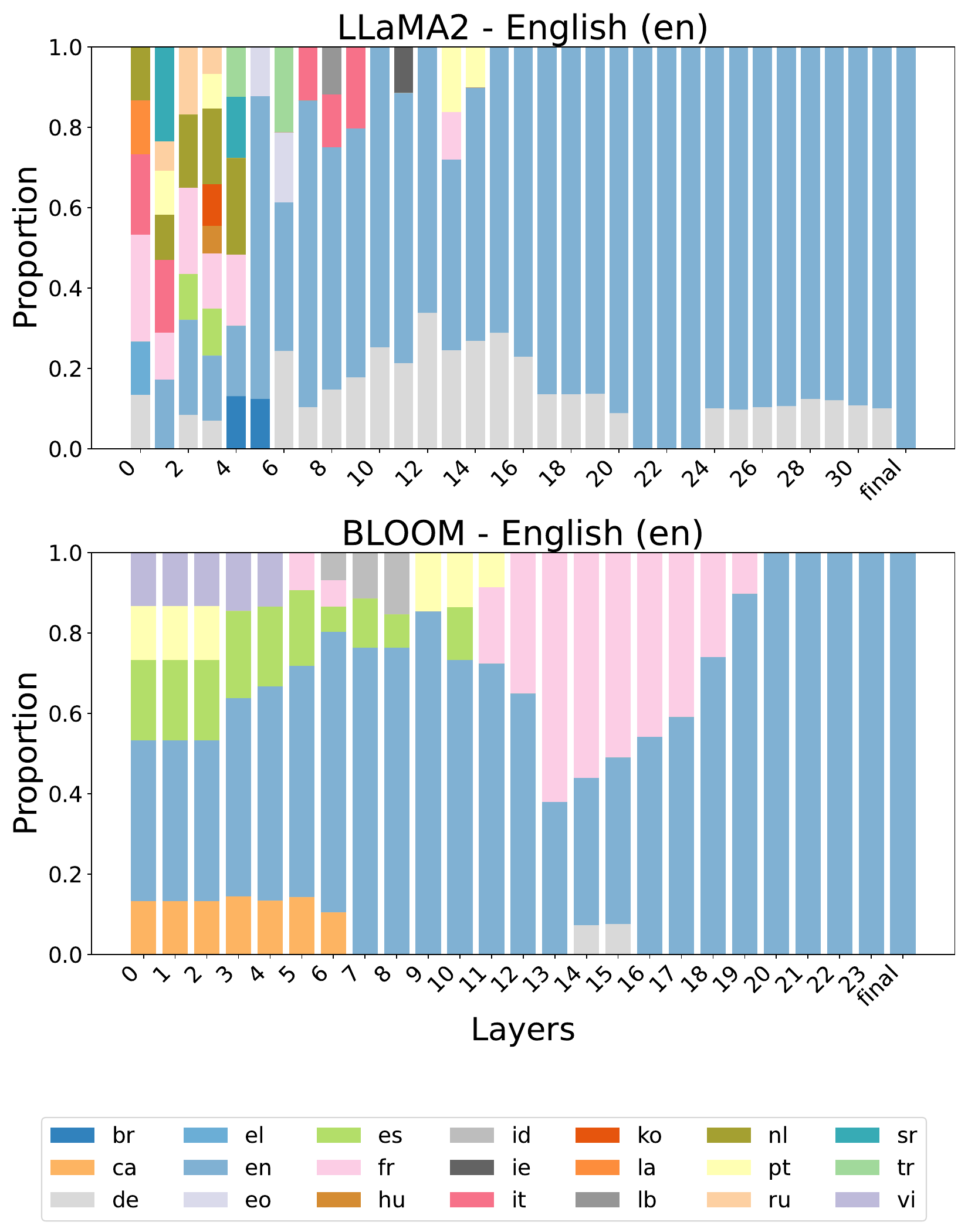}
    \caption{Language composition with English as the input language.}
    \label{fig:language-composition-en}
\end{subfigure}
\hfill
\begin{subfigure}{0.48\linewidth}
    \centering
    \includegraphics[width=\linewidth]{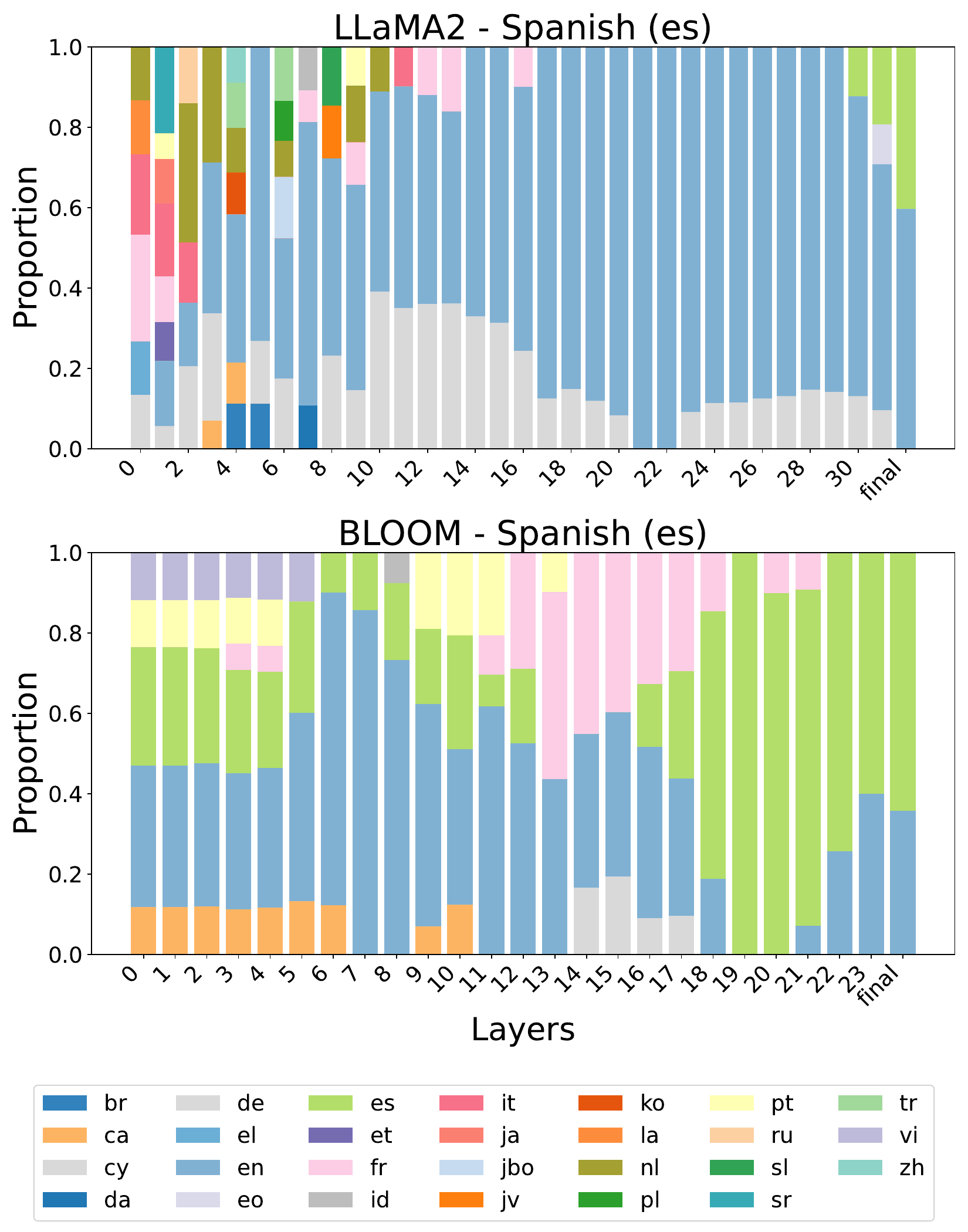}
    \caption{Language composition with Spanish as the input language.}
    \label{fig:language-composition-es}
\end{subfigure}
\end{figure*}
\begin{figure*}[h]
\ContinuedFloat
\begin{subfigure}{0.48\linewidth}
    \centering
    \includegraphics[width=\linewidth]{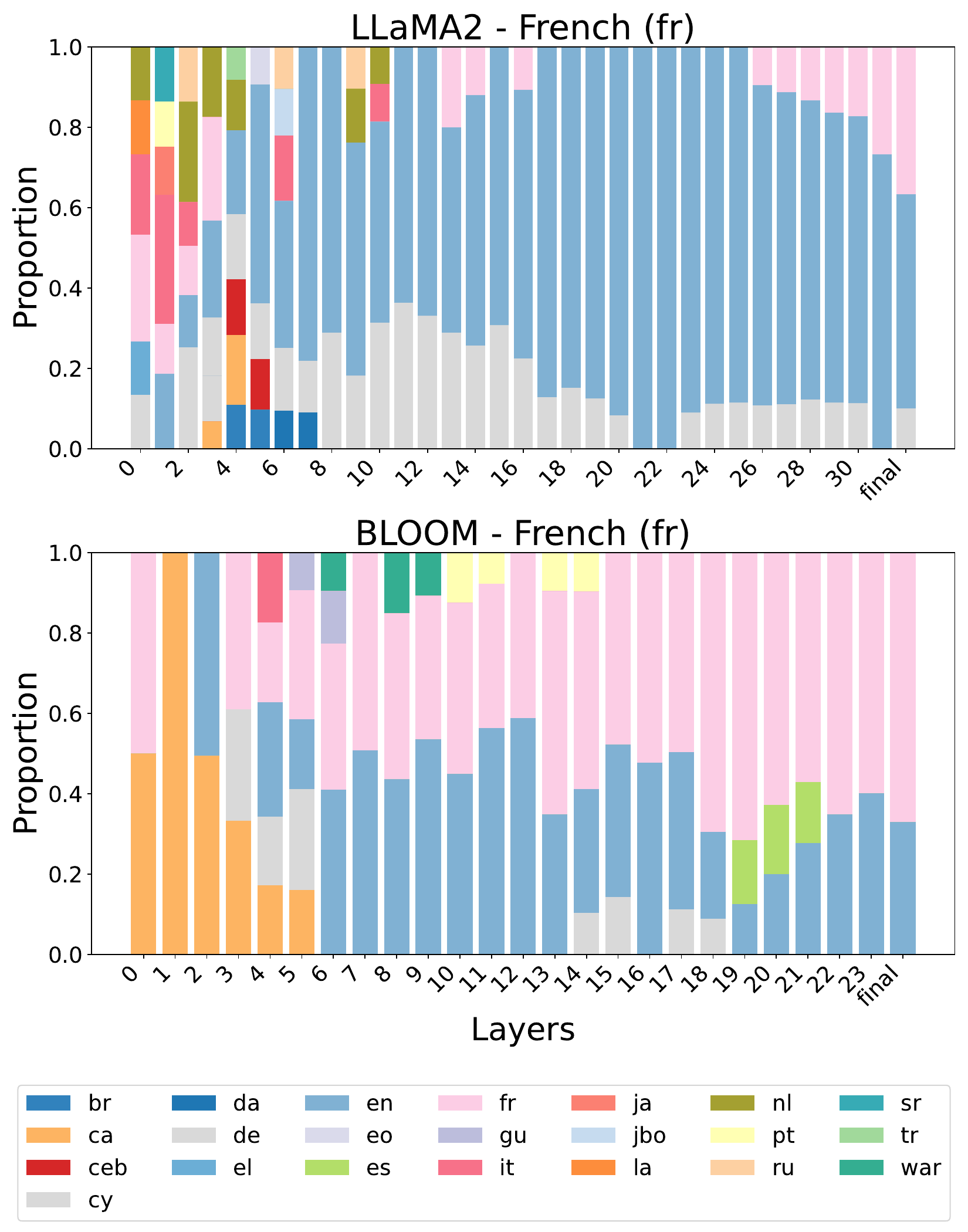}
    \caption{Language composition with French as the input language.\\}
    \label{fig:language-composition-fr}
\end{subfigure}
\hfill
\begin{subfigure}{0.48\linewidth}
    \centering
    \includegraphics[width=\linewidth]{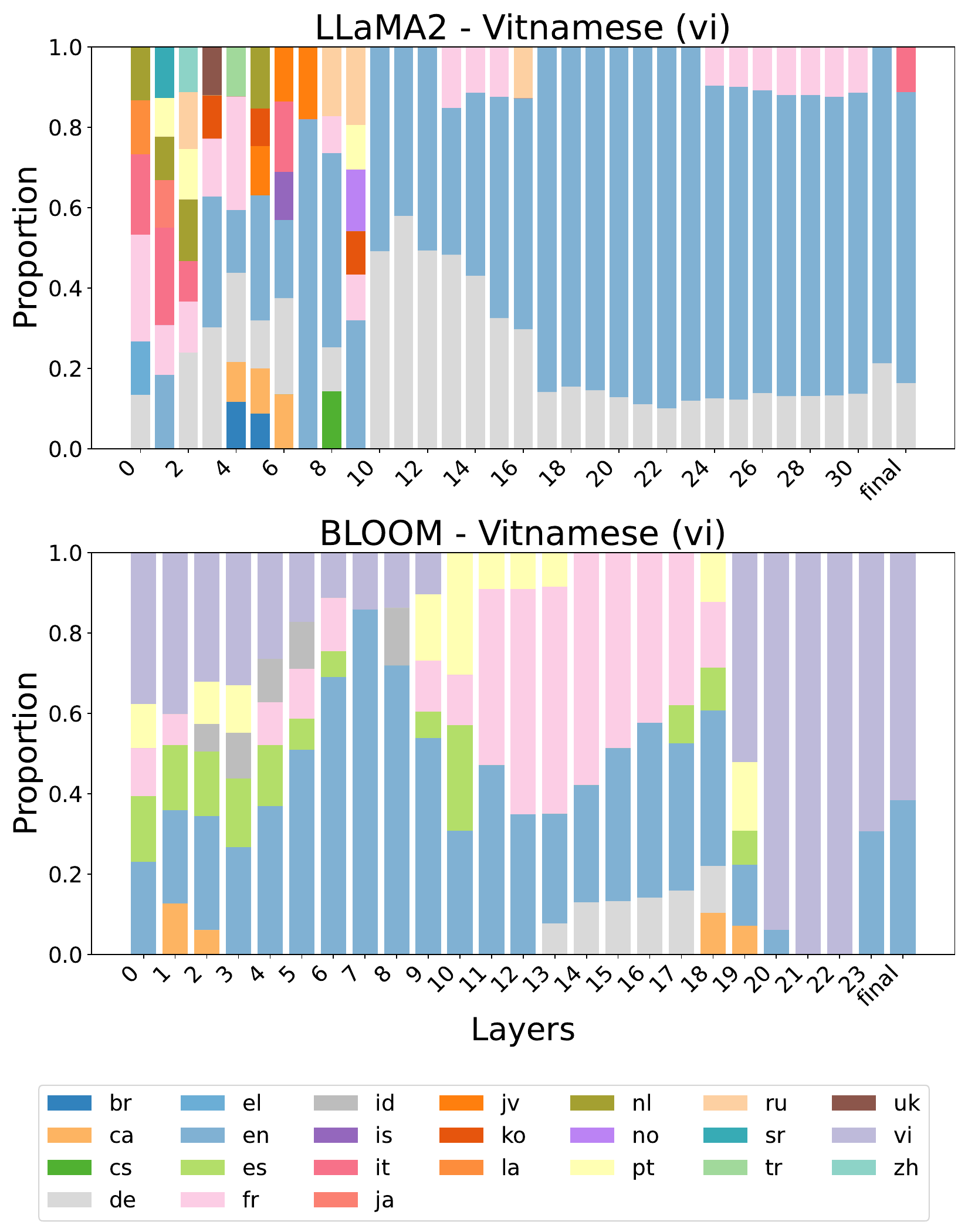}
    \caption{Language composition with Vietnamese as the input language.}
    \label{fig:language-composition-vi}
\end{subfigure}
\caption{Language composition for languages shared between LLaMA2 and BLOOM.}
\label{fig:language-composition-shared}
\end{figure*}

\begin{figure*}[h]
\begin{subfigure}{0.48\linewidth}
    \centering
    \includegraphics[width=\linewidth]{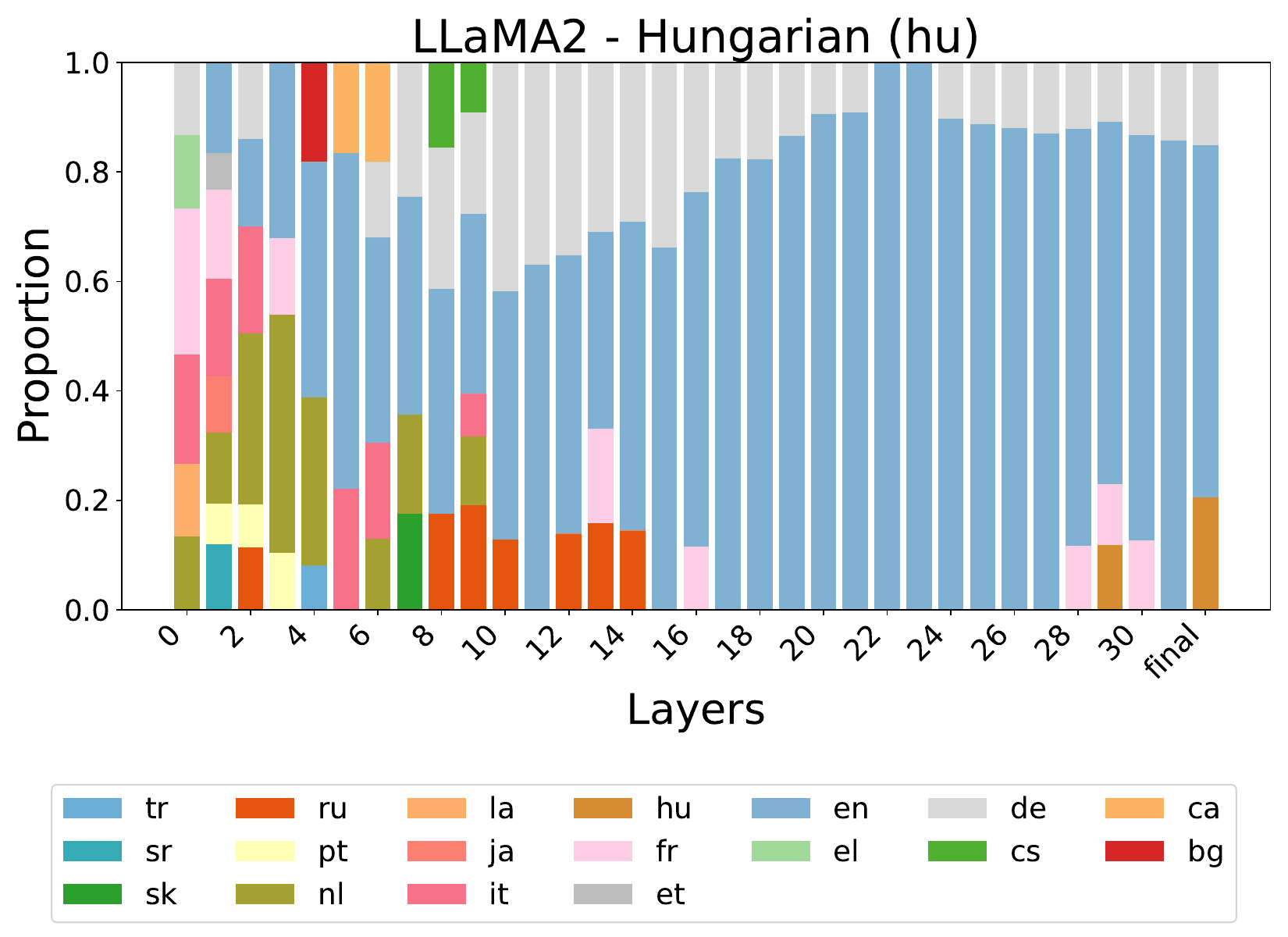}
    \caption{Language composition in LLaMA2 with Hungarian as the input language.}
    \label{fig:language-composition-hu}
\end{subfigure}
\hfill
\begin{subfigure}{0.48\linewidth}
    \centering
    \includegraphics[width=\linewidth]{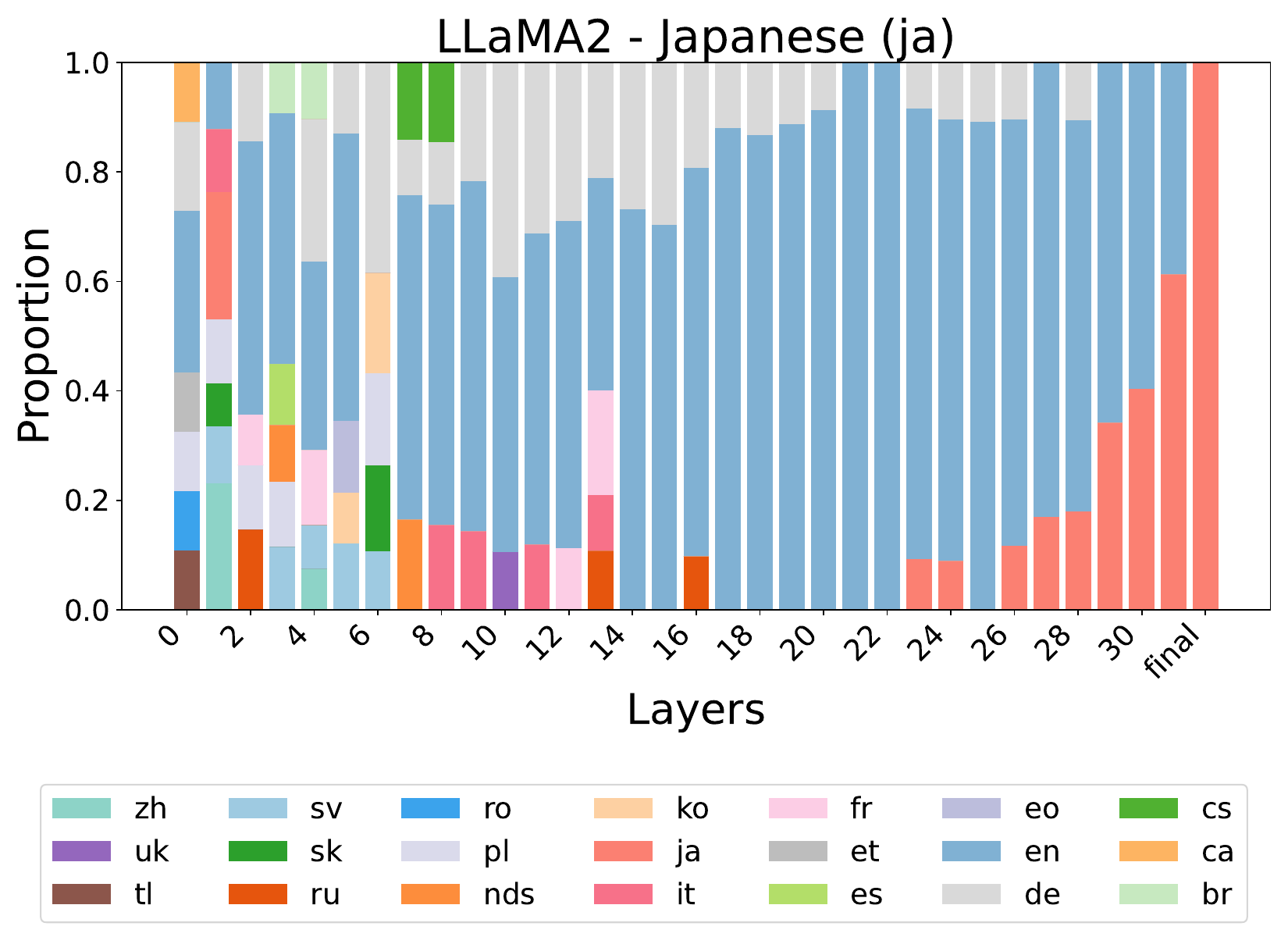}
    \caption{Language composition in LLaMA2 with Japanese as the input language.}
    \label{fig:language-composition-ja}
\end{subfigure}
\end{figure*}
\begin{figure*}[h]
\ContinuedFloat
\begin{subfigure}{0.48\linewidth}
    \centering
    \includegraphics[width=\linewidth]{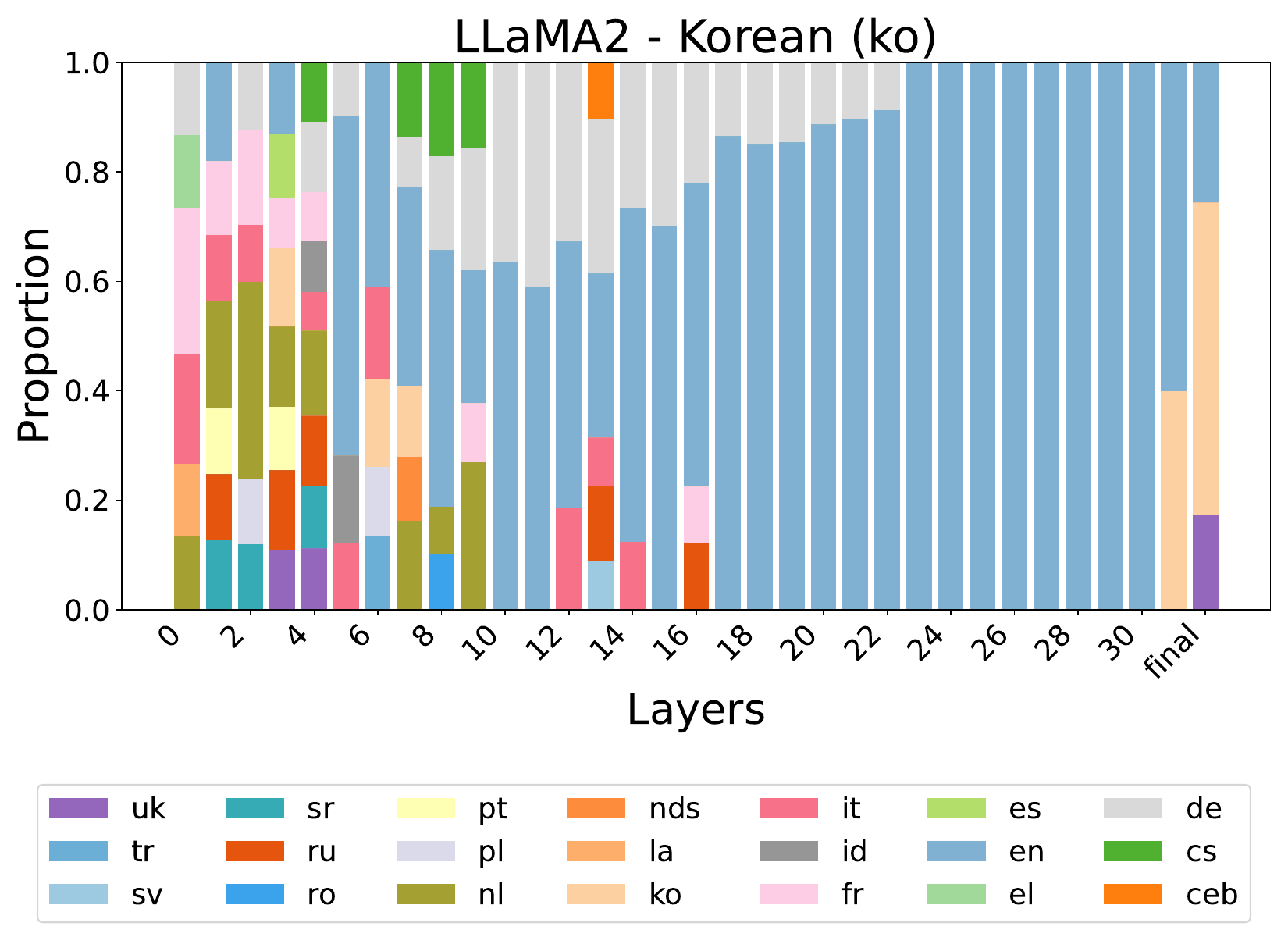}
    \caption{Language composition in LLaMA2 with Korean as the input language.}
    \label{fig:language-composition-ko}
\end{subfigure}
\hfill
\begin{subfigure}{0.48\linewidth}
    \centering
    \includegraphics[width=\linewidth]{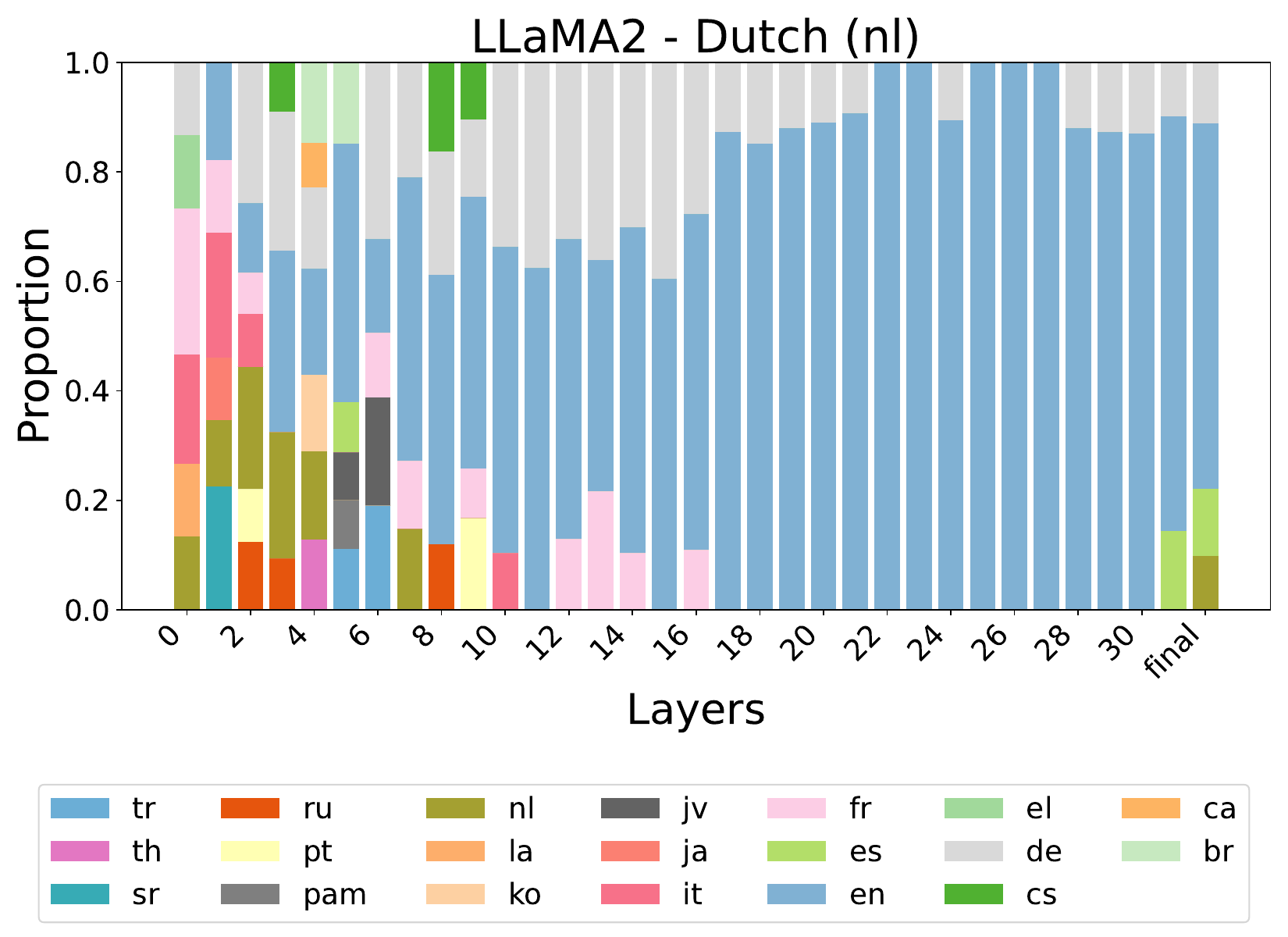}
    \caption{Language composition in LLaMA2 with Dutch as the input language.}
    \label{fig:language-composition-nl}
\end{subfigure}
\end{figure*}
\begin{figure*}[h]
\ContinuedFloat
\begin{subfigure}{0.48\linewidth}
    \centering
    \includegraphics[width=\linewidth]{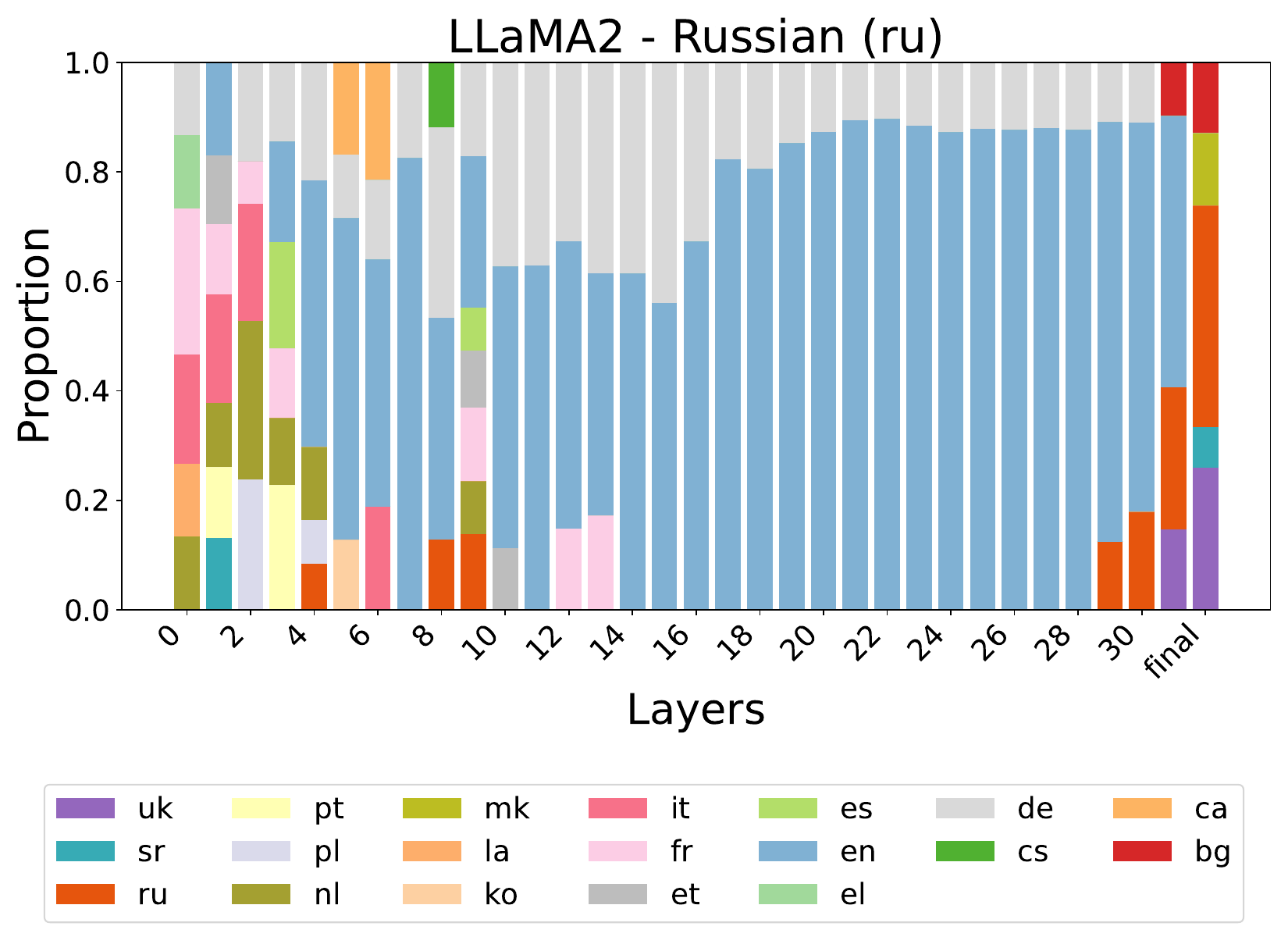}
    \caption{Language composition in LLaMA2 with Russian as the input language.}
    \label{fig:language-composition-ru}
\end{subfigure}
\hfill
\begin{subfigure}{0.48\linewidth}
    \centering
    \includegraphics[width=\linewidth]{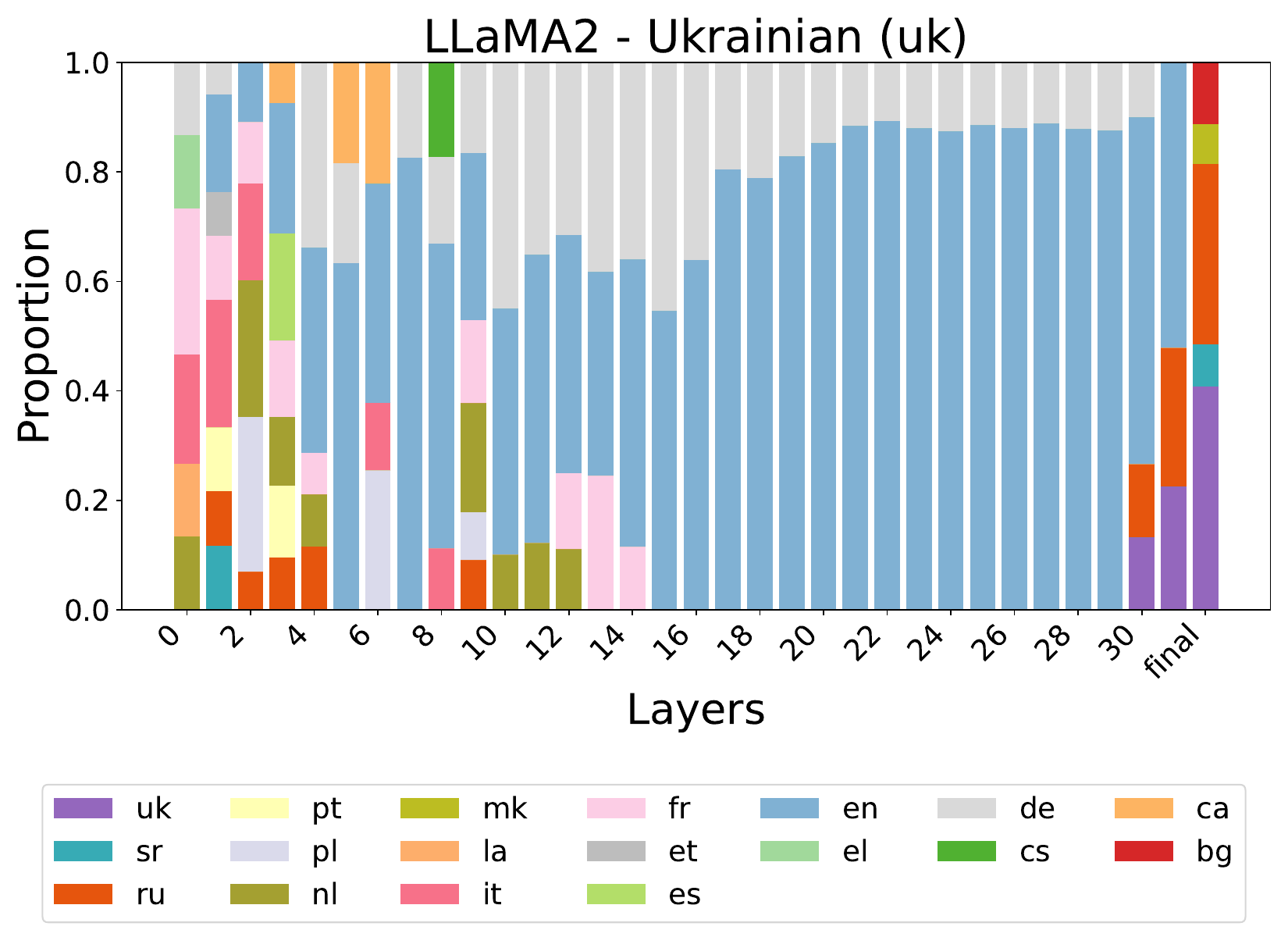}
    \caption{Language composition in LLaMA2 with Ukrainian as the input language.}
    \label{fig:language-composition-uk}
\end{subfigure}
\end{figure*}
\begin{figure*}[h]
\ContinuedFloat
\begin{subfigure}{\linewidth}
    \centering
    \includegraphics[width=0.48\linewidth]{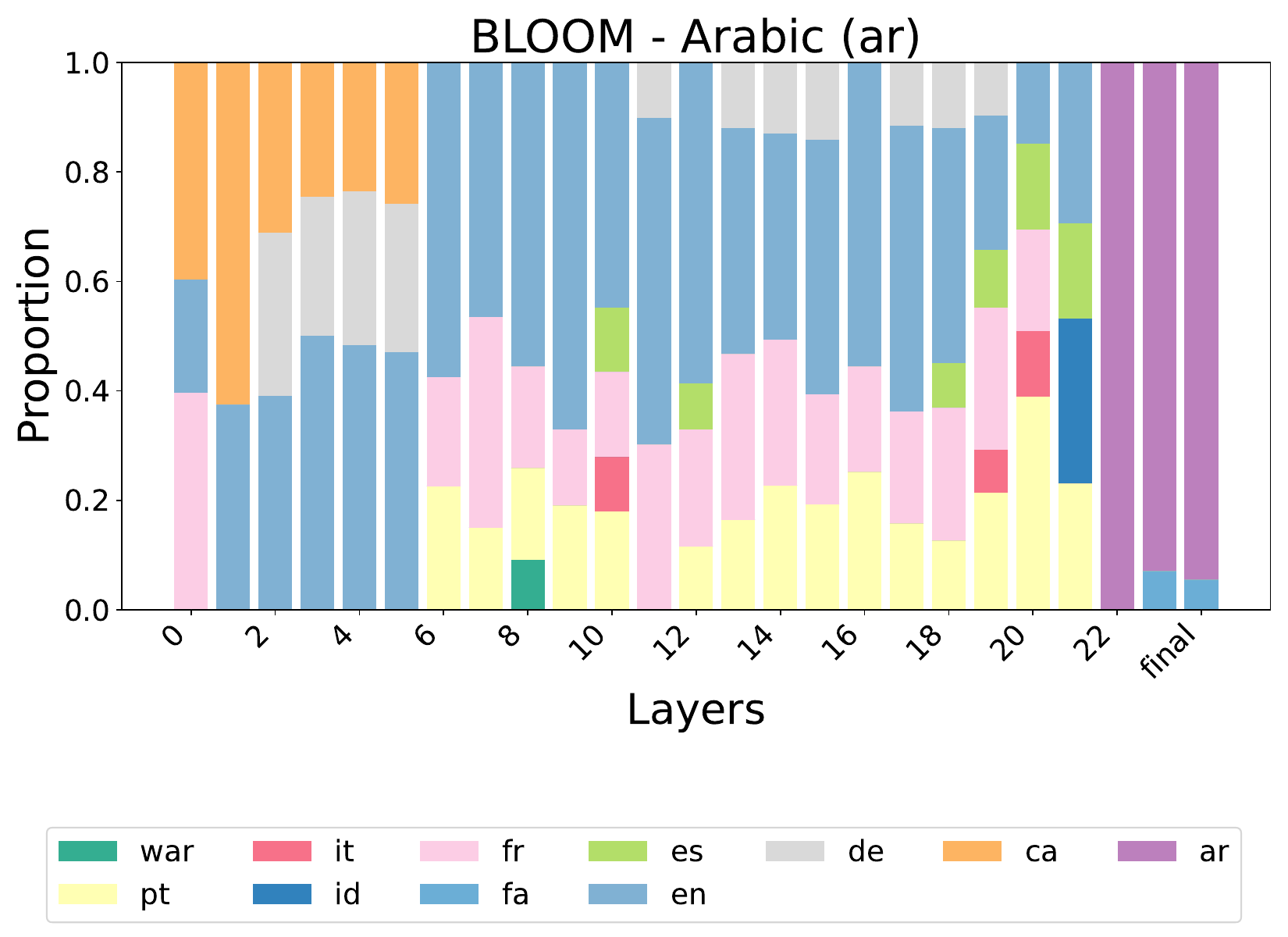}
    \caption{Language composition in BLOOM with Arabic as the input language.}
    \label{fig:language-composition-ar}
\end{subfigure}
\caption{Language composition for unique languages in LLaMA2 and BLOOM, respectively.}
\label{fig:language-composition-unique}
\end{figure*}

\begin{figure*}[!htbp]
    \centering
    \begin{subfigure}{0.48\linewidth}
    \includegraphics[width=\linewidth]{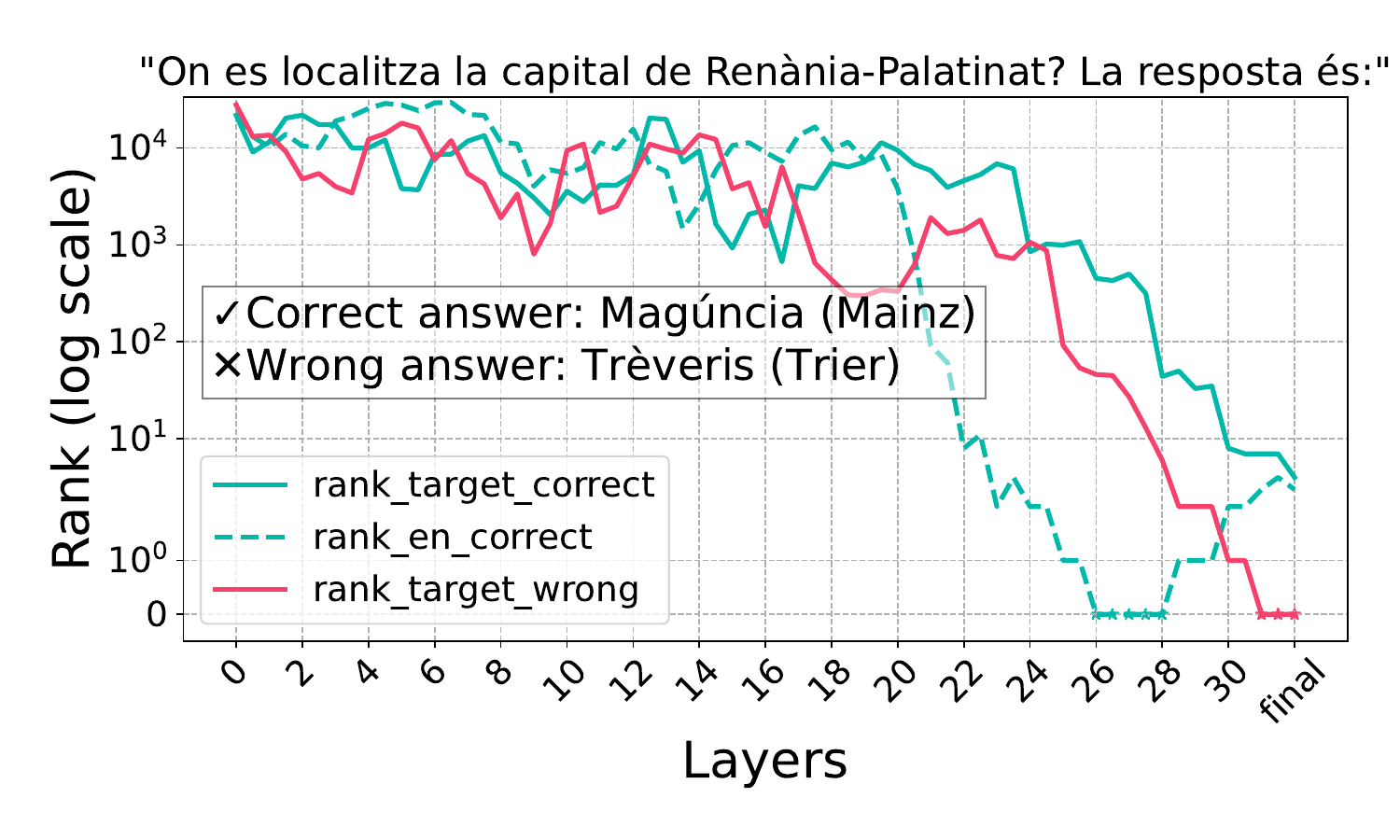}
        \caption{Prompt in Catalan; English translation: ``What is the capital of Rhineland-Palatinate? The answer is:''.}
        \label{fig:example-catalan}    
    \end{subfigure}
    \hfill
    \begin{subfigure}{0.48\linewidth}
    \includegraphics[width=\linewidth]{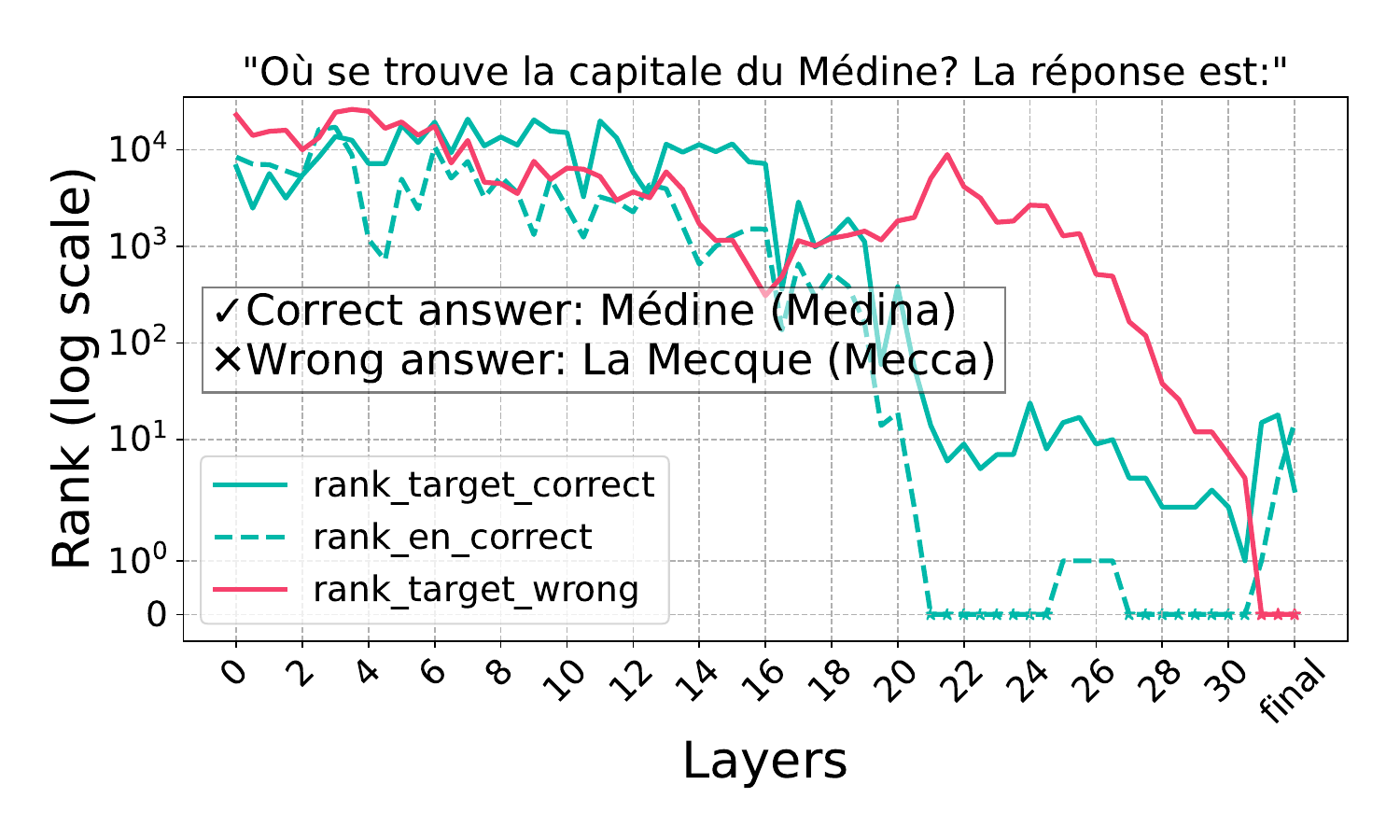}
        \caption{Prompt in French; English translation: ``What is the capital of Medina? The answer is:''. }
        \label{fig:example-french}    
    \end{subfigure}
\end{figure*}
\begin{figure*}[!htbp]
\ContinuedFloat
    \begin{subfigure}{0.48\linewidth}
    \includegraphics[width=\linewidth]{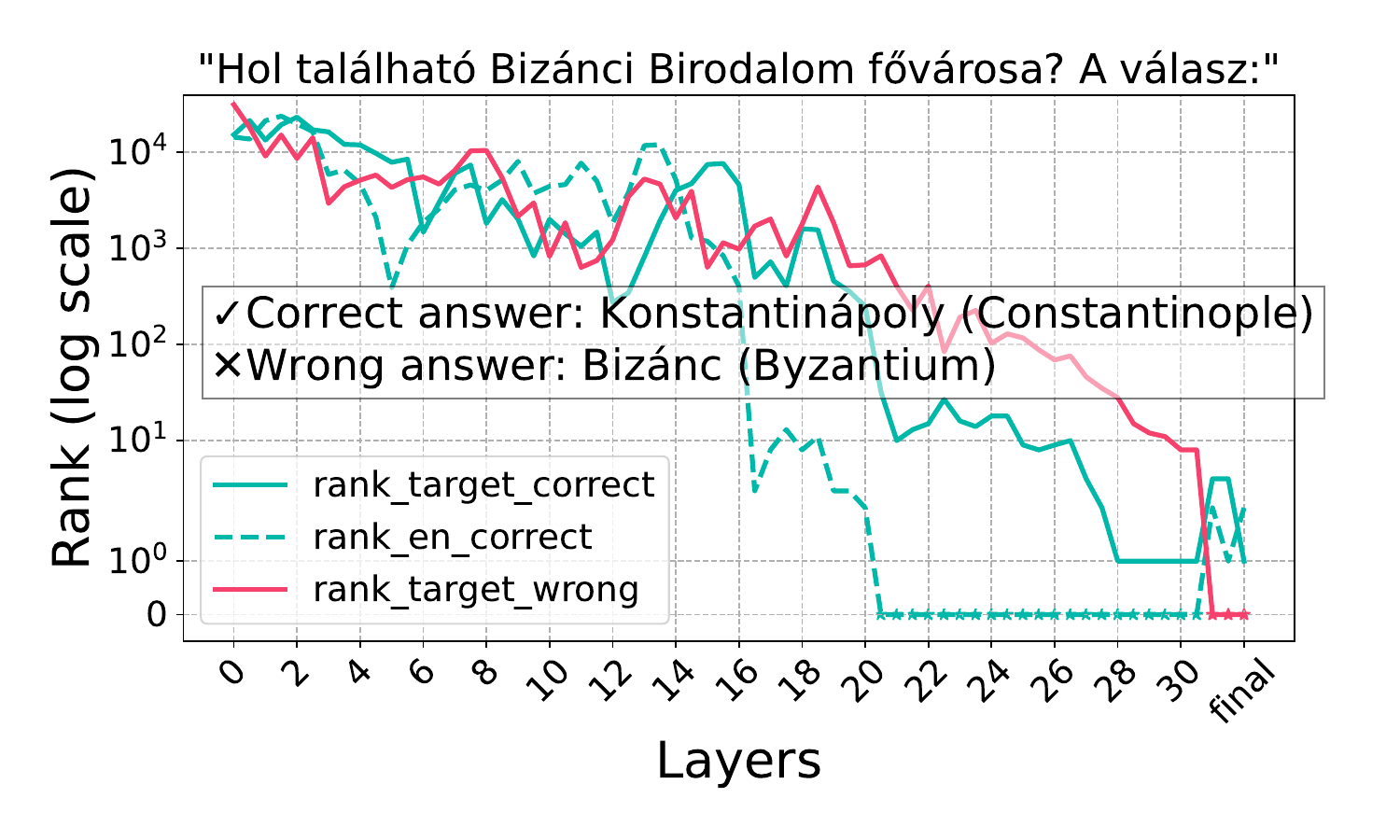}
        \caption{Prompt in Hungarian; English translation: ``What is the capital of Byzantine Empire? The answer is:''.}
        \label{fig:example-hungarian}    
    \end{subfigure}
    \hfill
    \begin{subfigure}{0.48\linewidth}
    \includegraphics[width=\linewidth]{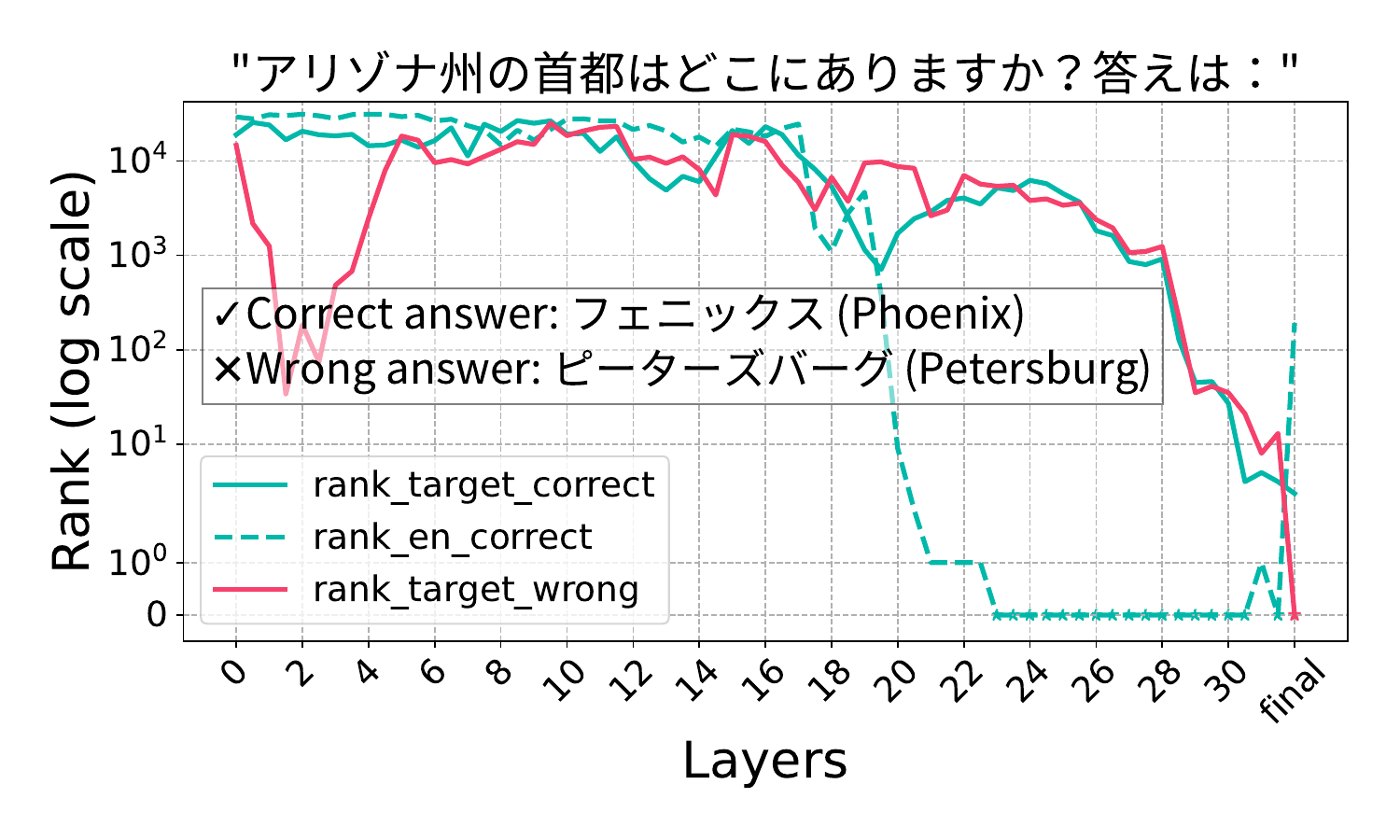}
        \caption{Prompt in Japanese; English translation: ``What is the capital of Arizona? The answer is:''.}
        \label{fig:example-japanese}
    \end{subfigure}
\end{figure*}
\begin{figure*}[!htbp]
\ContinuedFloat
    \begin{subfigure}{0.48\linewidth}
    \includegraphics[width=\linewidth]{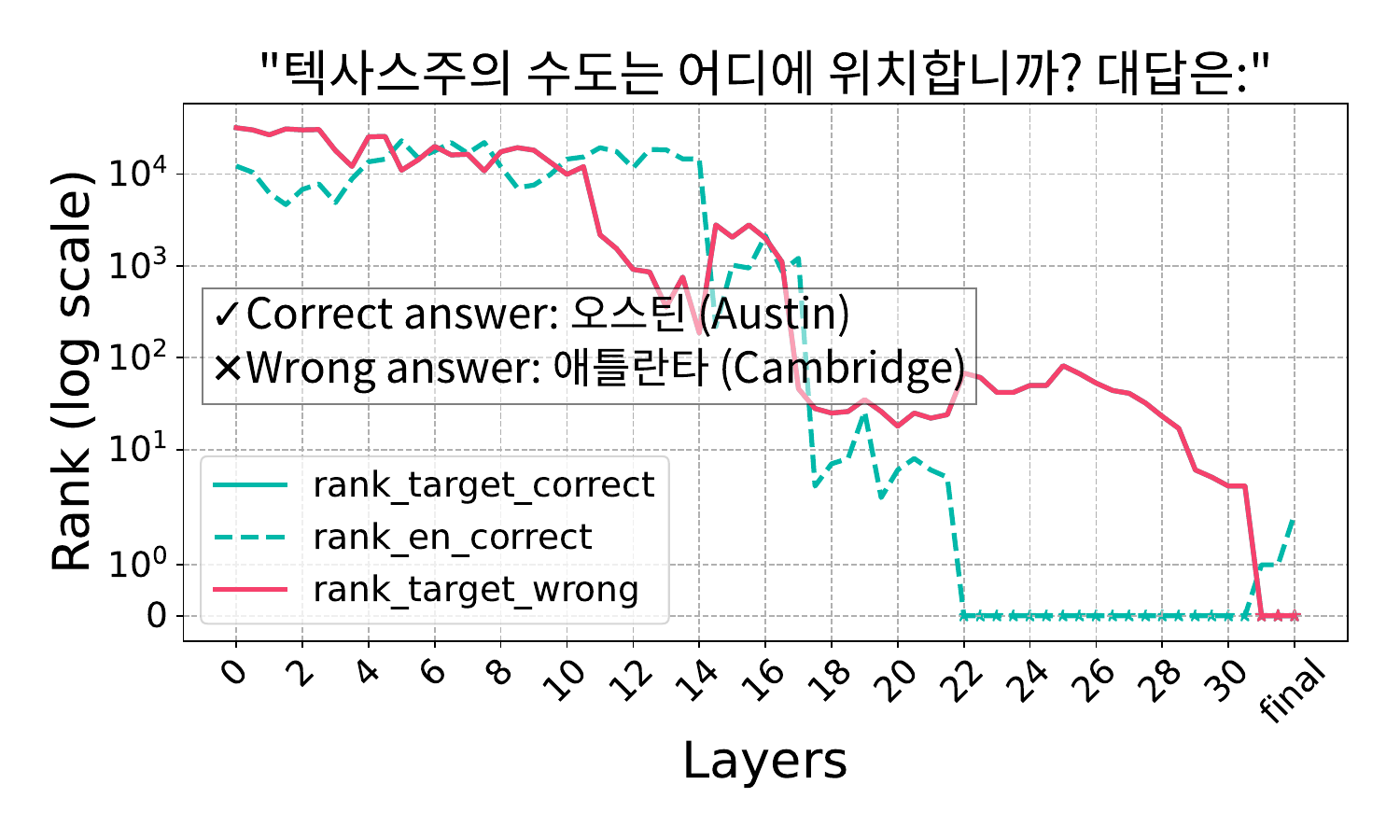}
        \caption{Prompt in Korean; English translation: ``What is the capital of Texas? The answer is:''.}
        \label{fig:example-korean}
    \end{subfigure}
    \hfill
    \begin{subfigure}{0.48\linewidth}
    \includegraphics[width=\linewidth]{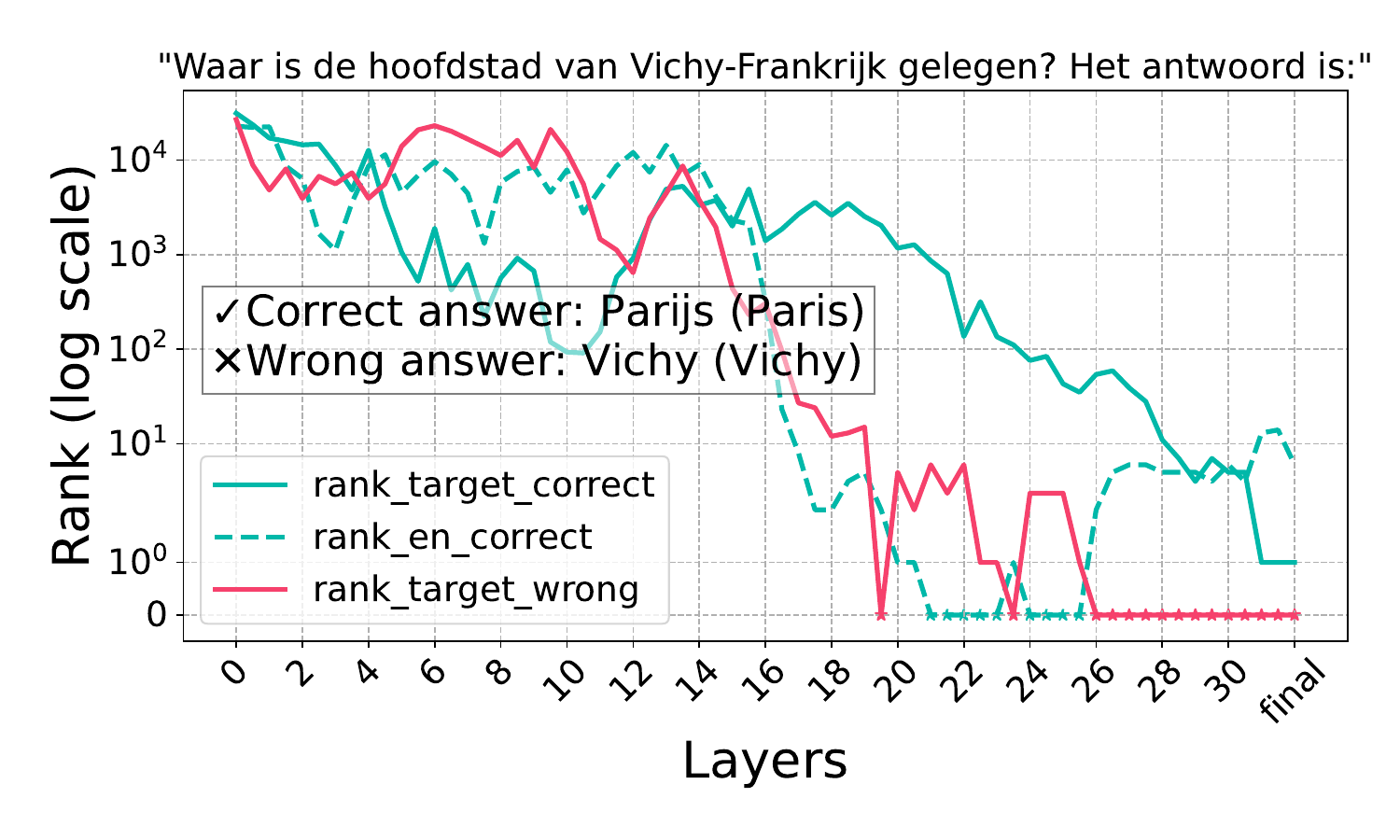}
        \caption{Prompt in Dutch; English translation: ``What is the capital of Vichy France? The answer is:''.}
        \label{fig:example-dutch}
    \end{subfigure}
\end{figure*}
\begin{figure*}[!htbp]
\ContinuedFloat
    \begin{subfigure}{0.48\linewidth}
        \includegraphics[width=\linewidth]{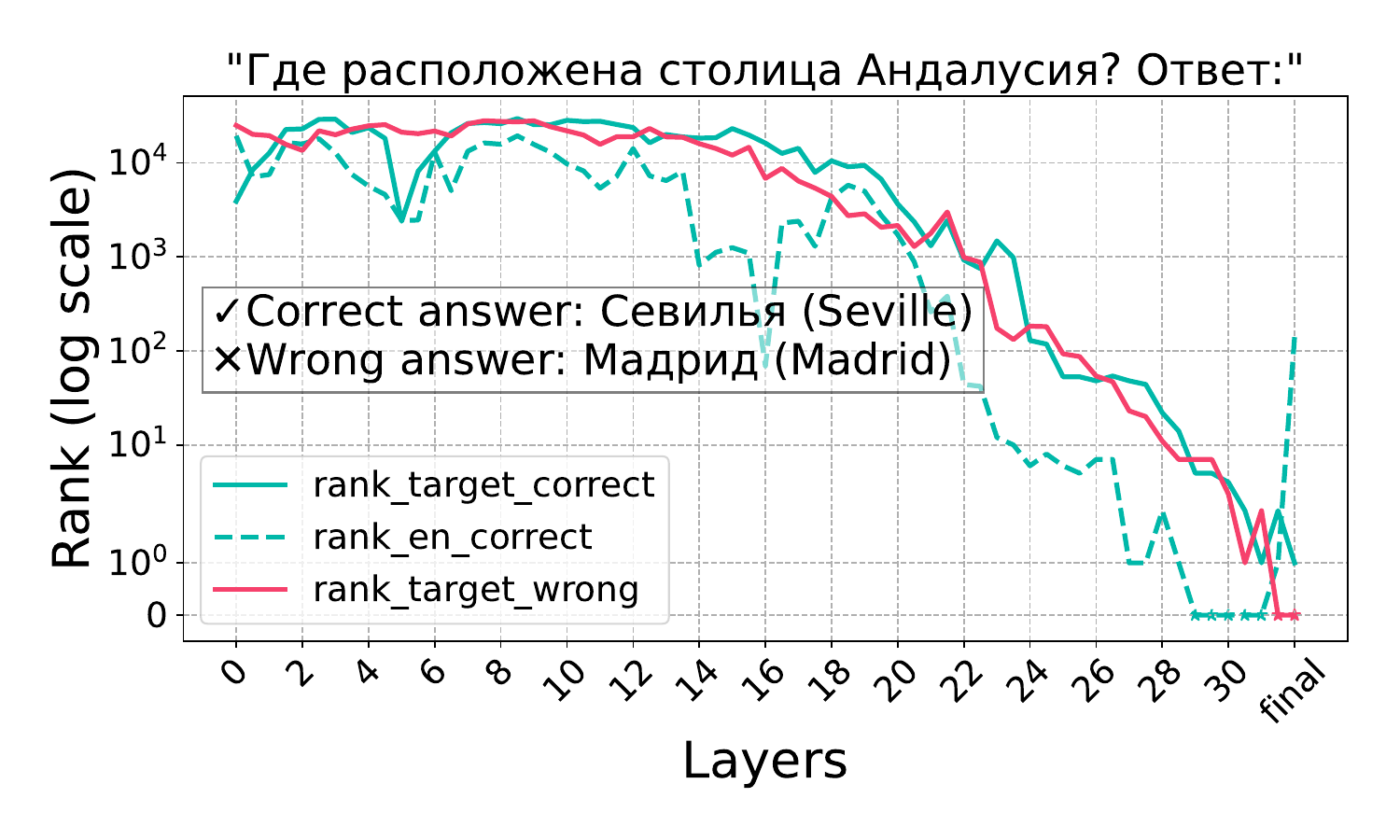}\caption{Prompt in Russian; English translation: ``What is the capital of Andalusia? The answer is:''.}
        \label{fig:example-russian}
    \end{subfigure}
    \hfill
    \begin{subfigure}{0.48\linewidth}
        \includegraphics[width=\linewidth]{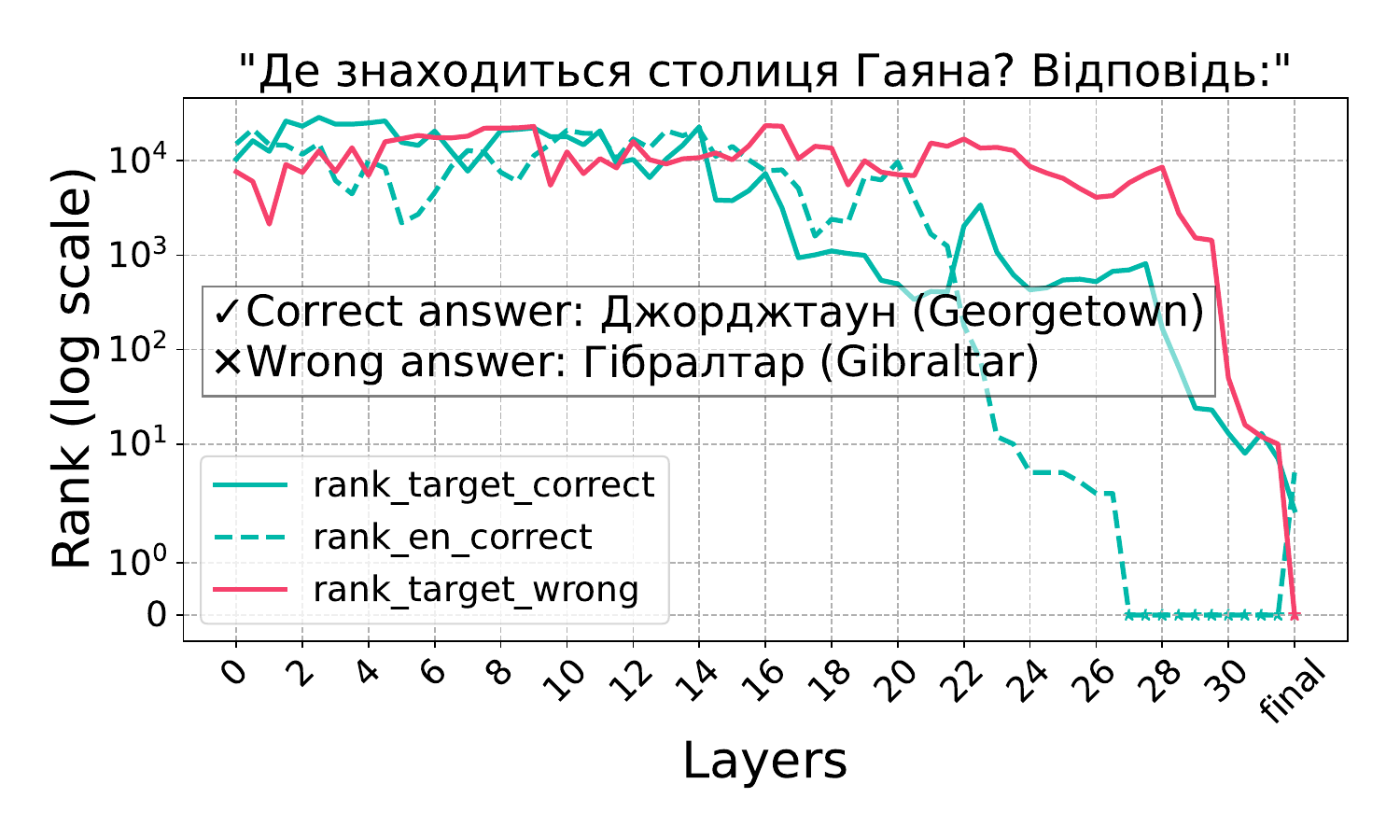}\caption{Prompt in Ukrainian; English translation: ``What is the capital of Guyana? The answer is:''.}
        \label{fig:example-ukrainian}
    \end{subfigure}
\end{figure*}
\begin{figure*}[!htbp]
\ContinuedFloat
\centering
    \begin{subfigure}{0.48\linewidth}
        \includegraphics[width=\linewidth]{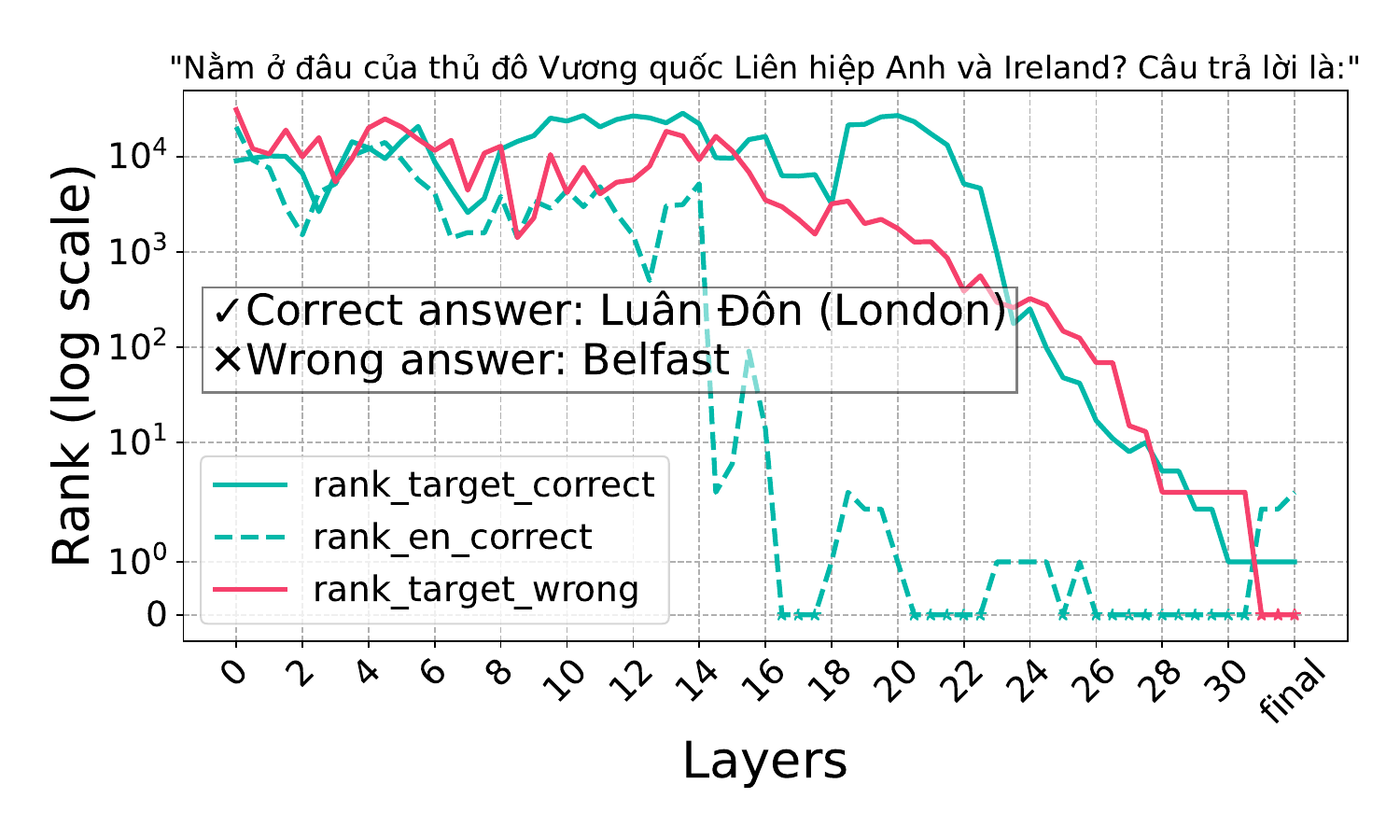}\caption{Prompt in Ukrainian; English translation: ``What is the capital of United Kingdom of Great Britain and Ireland? The answer is:''.}
        \label{fig:example-vietnamese}
    \end{subfigure}

    \caption{Rank evolution for prompts in different languages. \texttt{rank\_target\_wrong} represents the rank of the model's final incorrect prediction across layers, while \texttt{rank\_target\_correct} and \texttt{rank\_en\_correct} denote the ranks of the correct answer in the target language and the English equivalent, respectively. The plots show the impact of errors during language transition, where the rank of the incorrect answer surpasses the correct answer in the final layers.}
    \label{fig:app-rank-plot-examples}
\end{figure*}
\end{document}